%% file: 00-main.tex
\pdfoutput=1

\documentclass[11pt]{article}

\usepackage[]{emnlp2021}

\usepackage{times}
\usepackage{latexsym}

\usepackage[T1]{fontenc}

\usepackage[utf8]{inputenc}

\usepackage{microtype}

\usepackage{amsthm}
\usepackage{amssymb}
\usepackage{amsmath}
\usepackage{amsfonts}
\usepackage{xcolor}
\usepackage{multirow}
\usepackage{tabularx}
\usepackage{booktabs}
\usepackage{subcaption}
\usepackage{graphicx}
\usepackage{color,colortbl}
\usepackage{url}
\usepackage{xspace}
\usepackage{dsfont}
\usepackage{arydshln}
\usepackage{multirow}

\newcommand{\convfit}{{\textsc{ConvFiT}}\xspace}
\definecolor{Gray}{gray}{0.92}

\newcommand{\mneg}{{\textsc{mneg}}\xspace}
\newcommand{\nce}{{\textsc{ocl}}\xspace}
\newcommand{\softmax}{{\textsc{softmax}}\xspace}
\newcommand{\smax}{{\textsc{smax}}\xspace}
\newcommand{\cosine}{{\textsc{cos}}\xspace}

\newcommand{\banking}{{\textsc{banking77}}\xspace}
\newcommand{\hwu}{{\textsc{hwu64}}\xspace}
\newcommand{\clinc}{{\textsc{clinc150}}\xspace}

\newcommand{\rob}{{\textsc{rob}}\xspace}
\newcommand{\bert}{{\textsc{bert}}\xspace}
\newcommand{\drob}{{\textsc{drob}}\xspace}

\usepackage[disable]{todonotes}
\makeatletter
\newcommand*\iftodonotes{\if@todonotes@disabled\expandafter\@secondoftwo\else\expandafter\@firstoftwo\fi}
\makeatother

\title{\convfit: Conversational Fine-Tuning of Pretrained Language Models}

\author{
Ivan Vuli\'{c}, Pei-Hao Su, Sam Coope, Daniela Gerz, \\
{\bf Pawe\l{} Budzianowski, I\~{n}igo Casanueva, Nikola Mrk\v{s}i\'{c}, \textnormal{and} Tsung-Hsien Wen}
 \vspace{2mm} \\
 PolyAI Limited \\
 London, United Kingdom \\
 \texttt{www.polyai.com}
}

\begin{document}
\maketitle
\begin{abstract}
Transformer-based language models (LMs) pretrained on large text collections are proven to store a wealth of semantic knowledge. However, \textbf{1)} they are not effective as sentence encoders when used off-the-shelf, and \textbf{2)} thus typically lag behind conversationally pretrained (e.g., via response selection) encoders on conversational tasks such as intent detection (ID). In this work, we propose \convfit, a simple and efficient two-stage procedure which turns any pretrained LM into a universal conversational encoder (after Stage 1 \convfit-ing) and task-specialised sentence encoder (after Stage 2). We demonstrate that \textbf{1)} full-blown conversational pretraining is not required, and that LMs can be quickly transformed into effective conversational encoders with much smaller amounts of unannotated data; \textbf{2)} pretrained LMs can be fine-tuned into task-specialised sentence encoders, optimised for the fine-grained semantics of a particular task. Consequently, such specialised sentence encoders allow for treating ID as a simple semantic similarity task based on interpretable nearest neighbours retrieval. We validate the robustness and versatility of the \convfit framework with such similarity-based inference on the standard ID evaluation sets: \convfit-ed LMs achieve state-of-the-art ID performance across the board, with particular gains in the most challenging, few-shot setups.
\end{abstract}

\section{Introduction and Motivation}
\label{s:intro}
\input{01-intro}

\section{Methodology}
\label{s:methodology}
\input{02-methodology}

\section{Experimental Setup}
\label{s:experimental}
\input{03-exp}

\section{Results and Discussion}
\label{s:results}
\input{04-results}

\section{Conclusion and Future Work}
\label{s:conclusion}
\input{05-conclusion}

\section*{Acknowledgements}
We are grateful to our colleagues at PolyAI for many fruitful discussions. We also thank the anonymous reviewers for their helpful suggestions.

\clearpage
\bibliography{refs}
\bibliographystyle{acl_natbib}

\clearpage
\appendix
\input{xx-appendix.tex}

\end{document}

%% file: 01-intro.tex
Pretrained Transformer-based (masked) language models (LMs) such as BERT \cite{Devlin:2018arxiv} or RoBERTa \cite{Liu:2019roberta}, coupled with task-specific fine-tuning, offer unmatched state-of-the-art performance in a wide array of standard language understanding and conversational tasks \cite{Wang:2019superglue,Mehri:2020dialoglue}. However, pretrained LMs do not produce coherent and effective sentence encodings off-the-shelf; their further adaptation is required, akin to standard task fine-tuning. For instance, \newcite{Reimers:2019emnlp} transform monolingual English BERT with supervised natural language inference and paraphrasing data \cite{Williams:2018naacl,Wieting:2018acl} into a sentence encoder which excels at sentence similarity and retrieval tasks \cite{Marelli:2014lrec,Cer:2017semeval}. This transformation process supports the creation of other similar universal sentence encoders in monolingual and multilingual settings \cite{Chidambaram:2019repl,Wieting:2020emnlp,Feng:2020labse}, and is typically based on dual-encoder architectures.

Another parallel research thread aims at learning \textit{conversational} encoders: it validates the benefits of masked language modeling (MLM) pretraining on naturally conversational data \cite{Wu:2020tod,Mehri:2021naacl}, as well as the benefits of transfer learning for conversational tasks which goes beyond MLM as the pretraining objective \cite[\textit{inter alia}]{Mehri:2019acl,CoopeFarghly2020,Henderson:2021convex}. In particular, response selection as a suitable pretraining task \cite{AlRfou:2016arxiv,Yang:2018repl,Henderson:2019acl,Humeau:2019arxiv} learns representations that organically capture conversational cues from conversational text data such as Reddit \cite{Henderson:2019arxiv}, again via dual-encoder architectures.

\begin{figure*}[t]
    \centering
    \includegraphics[width=0.97\textwidth]{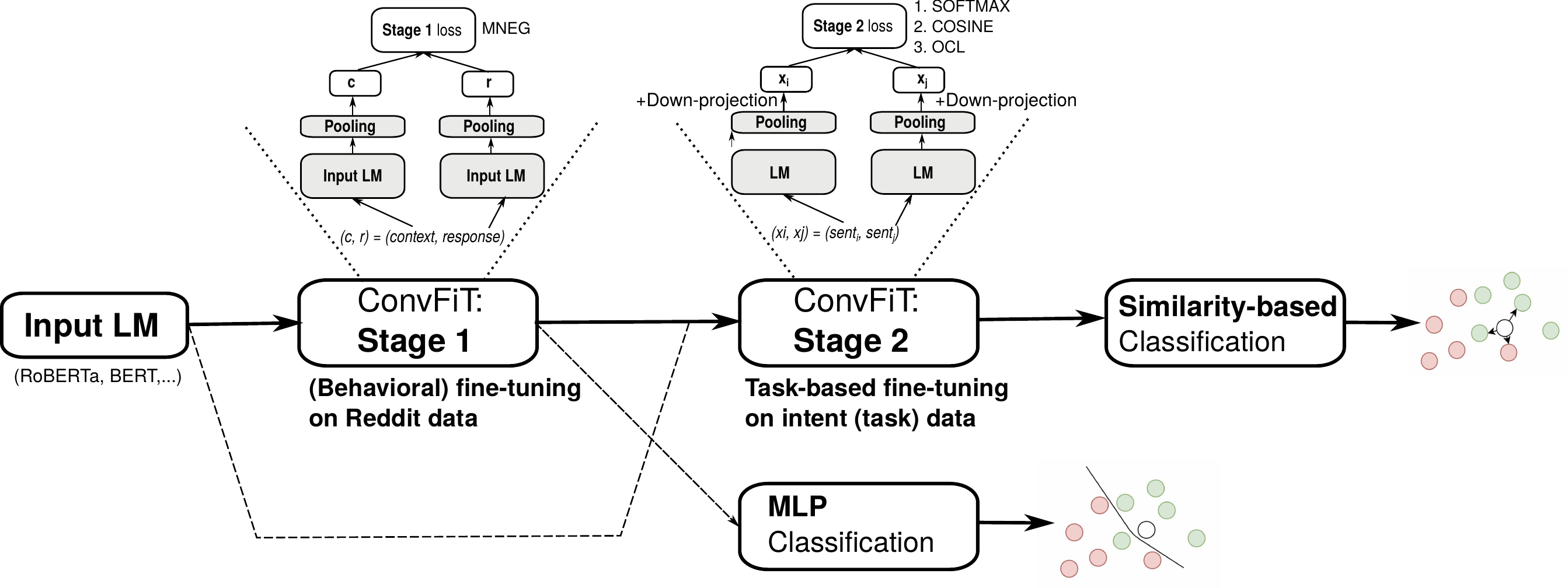}
    \vspace{-0.5mm}
    \caption{Illustration of the full \convfit framework which fine-tunes pretrained LMs such as BERT or RoBERTa in two separate stages via dual-encoder networks (``zoomed-in'' parts; grey blocks denote tunable parameters), and performs intent detection with the \convfit-ed models via similarity-based inference. \textbf{Stage 1 (S1):} adaptive conversational fine-tuning, \S\ref{ss:stage1}; \textbf{Stage 2 (S2):} task-tailored conversational fine-tuning (for intent detection), \S\ref{ss:stage2}. Dashed lines denote baseline/ablation variants which skip one of the two stages: (i) we can directly task-tune the sentence encoder with the task data (Stage 2) without running Stage 1, or (ii) we can skip Stage 2, and similar to \newcite{Casanueva:2020ws}, learn an MLP classifier on top of the conversational representations from Stage 1.}
    \vspace{-0.5mm}
    \label{fig:overview}
\end{figure*}
\begin{figure*}[t]
    \centering
    \begin{subfigure}[!ht]{0.97\textwidth}
        \centering
        \includegraphics[width=1.0\linewidth,trim=0cm 0cm 0cm 0.1cm,clip]{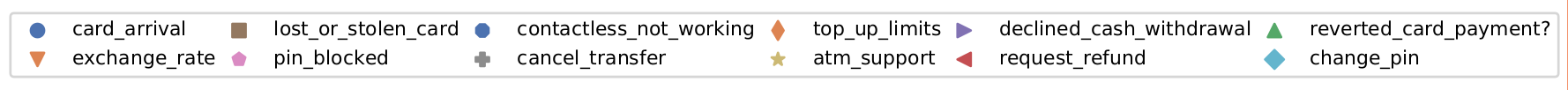}
        \label{fig:legend}
        \vspace{-3.5mm}
    \end{subfigure}
    \begin{subfigure}[!ht]{0.303\linewidth}
        \centering
        \includegraphics[width=0.98\linewidth]{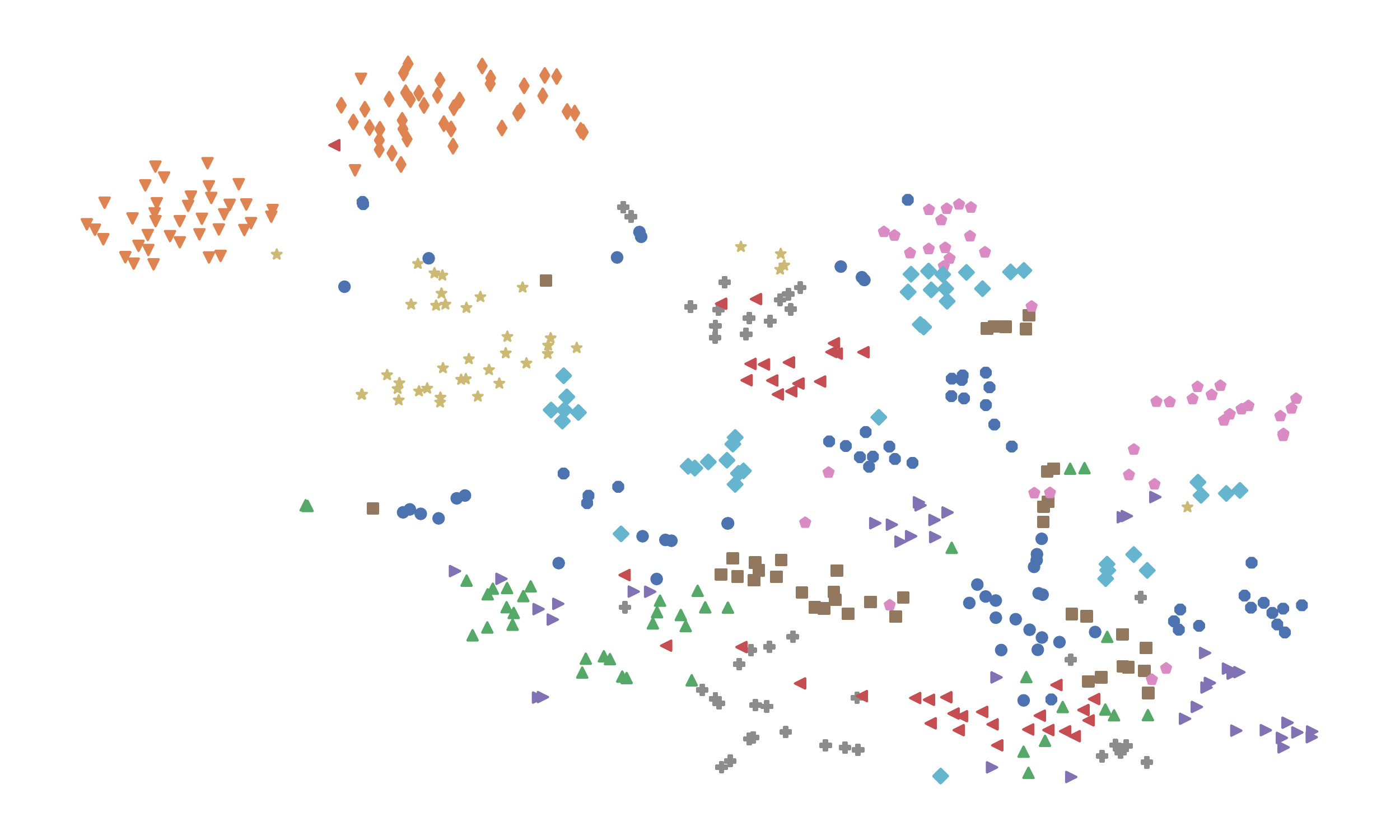}
        \caption{RoBERTa (no fine-tuning)}
        \label{fig:tsne-orig}
    \end{subfigure}
    \begin{subfigure}[!ht]{0.303\textwidth}
        \centering
        \includegraphics[width=0.98\linewidth]{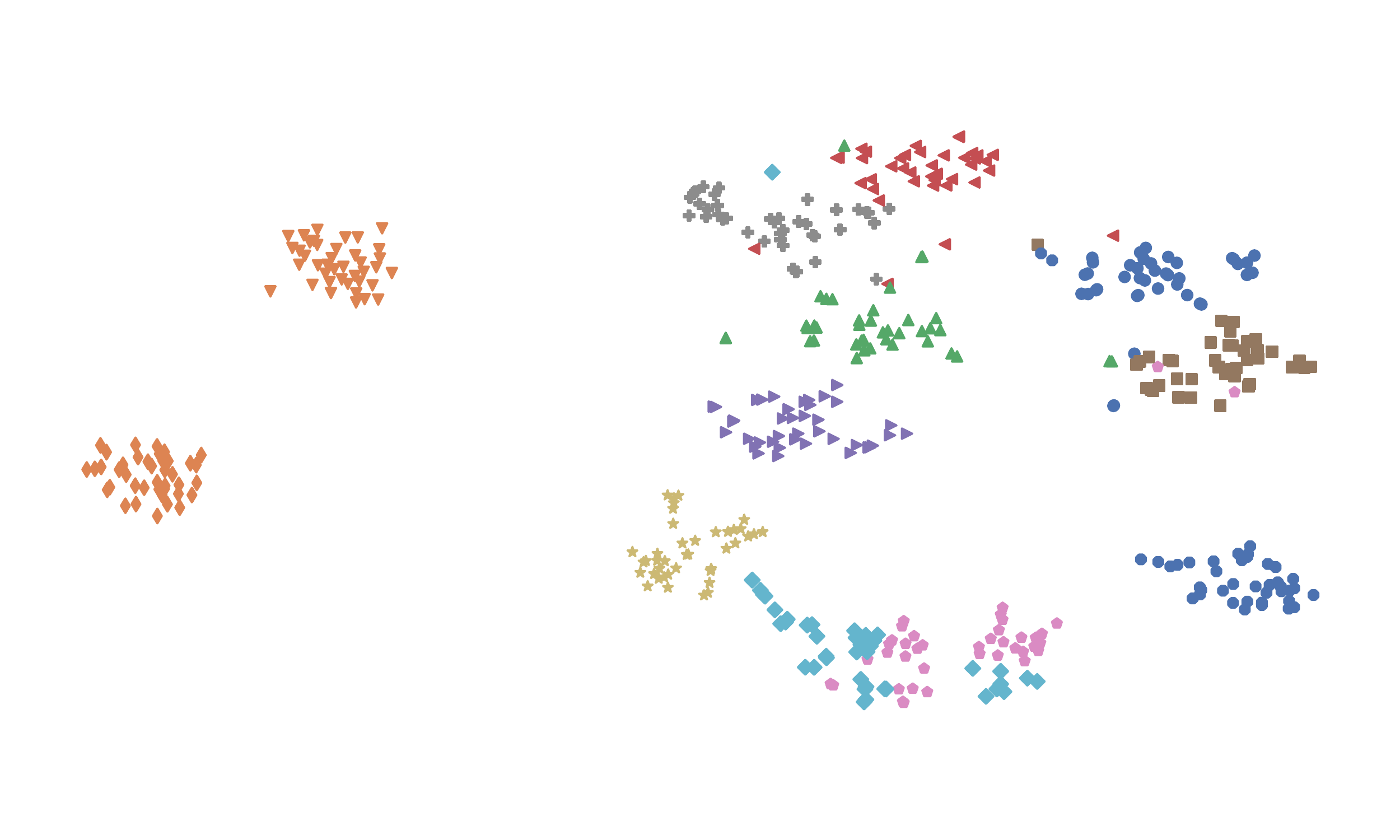}
        \caption{RoBERTa (after S1)}
        \label{fig:tsne-r15m}
    \end{subfigure}
    \begin{subfigure}[!ht]{0.303\linewidth}
        \centering
        \includegraphics[width=0.98\linewidth]{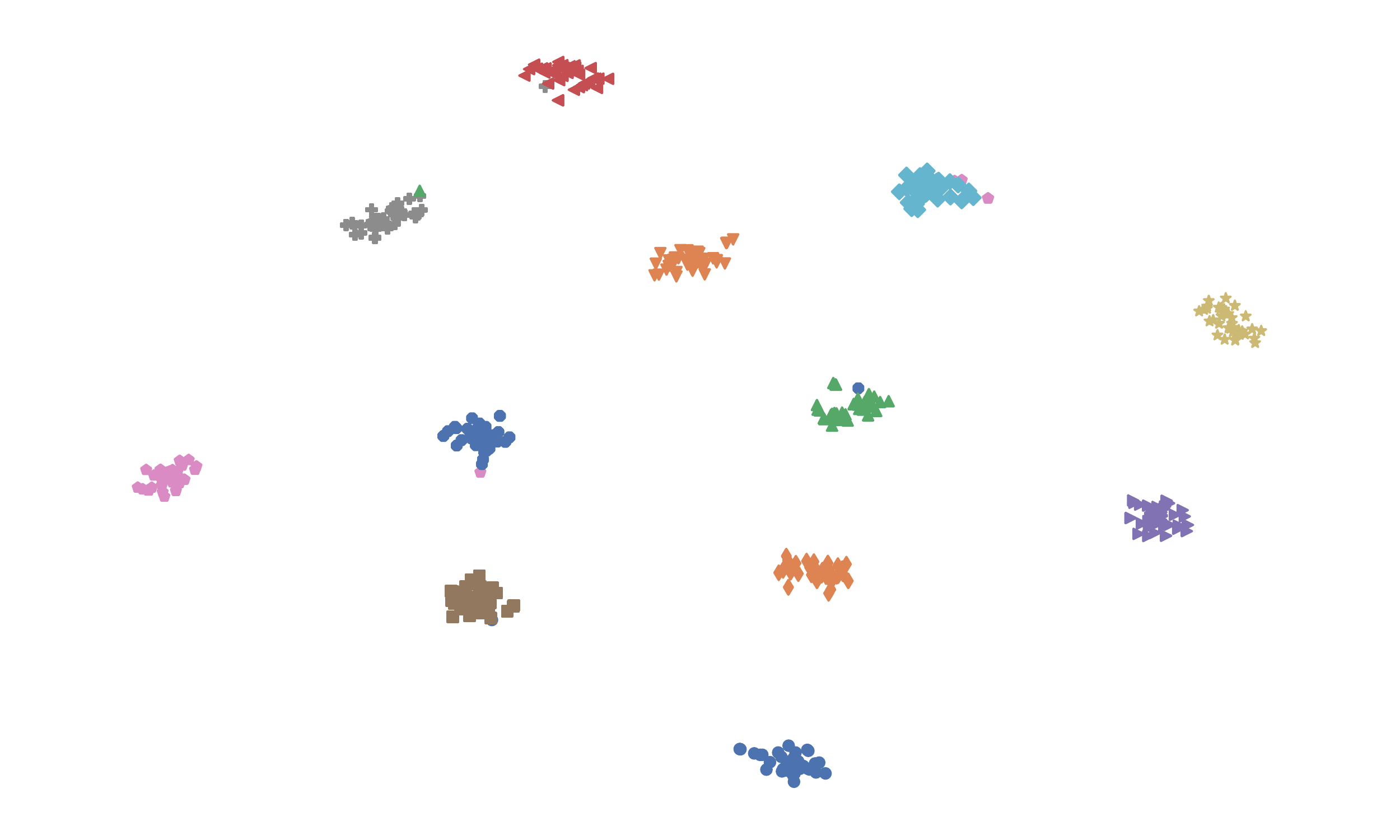}
        \caption{RoBERTa (after S1 and S2)}
        \label{fig:tsne-ocl}
    \end{subfigure}
    \vspace{-0.5mm}
    \caption{t-SNE plots \cite{tsne:2012} of encoded utterances from the ID test set of \textsc{banking77} (i.e., all examples are effectively unseen by the encoder models at training) associated with a selection of 12 intents, demonstrating the effects of gradual ``representation specialisation funnel''. The encoded utterances are created via mean-pooling based on \textbf{(a)} the original RoBERTa LM; \textbf{(b)} RoBERTa after Stage 1 (i.e., fine-tuned on 1\% of the full Reddit corpus, see Figure~\ref{fig:overview}); \textbf{(c)} RoBERTa after Stage 1 and Stage 2, fine-tuned with the \nce objective ($n=3$ negatives) using the entire \banking training set (see Figure~\ref{fig:overview}). Additional t-SNE plots are in the Appendix.}
    \vspace{-0.5mm}
\label{fig:tsne-rob}
\end{figure*}

Inspired by these two research threads, we pose the following two crucial questions:

\vspace{0.7mm}
\indent \textbf{(Q1)} Is it necessary to conduct full-scale expensive conversational pretraining? In other words, is it possible to simply and quickly 'rewire' existing MLM-pretrained encoders as conversational encoders via, e.g., response ranking fine-tuning on (much) smaller-scale datasets?

\vspace{0.7mm}
\indent \textbf{(Q2)} If we frame conversational tasks such as intent detection as semantic similarity tasks instead of their standard classification-based formulation, is it also possible to frame supervised task-specific learning as fine-tuning of conversational sentence encoders? In other words, can we learn task-specialised sentence encoders that enable \textit{sentence similarity-based} interpretable classification?




In order to address these two questions, we propose \textbf{\convfit}, a two-stage \textbf{\textsc{Conv}}ersational \textbf{\textsc{Fi}}ne-\textbf{\textsc{T}}uning procedure that turns general-purpose MLM-pretrained encoders into sentence encoders specialised for a particular conversational domain and task. Casting the end-task (e.g., intent detection) as a pure sentence similarity problem then allows us to recast task-tailored fine-tuning of a pretrained LM as gradual \textit{sentence encoder specialisation}, as illustrated in Figures~\ref{fig:overview} and \ref{fig:tsne-rob}.

Our hypothesis is that the pretrained LMs, which already store a wealth of semantic knowledge, can be gradually turned into conversational task-adapted sentence encoders without expensive full pretraining. \textbf{(S1)} Stage 1 transforms pretrained LMs into universal conversational encoders via \textit{adaptive fine-tuning} \cite{Ruder:2021blog} on (a fraction of) Reddit data (see Figure~\ref{fig:tsne-r15m}), relying on a standard dual-encoder architecture with a conversational response ranking loss \cite{Henderson:2020convert}; cf.~Q1. \textbf{\textbf{(S2)}} Stage 2 further specializes the sentence encoder via \textit{contrastive learning with in-task data}, that is, it learns meaningful task-related semantic clusters/subspaces. We then show that the S2 task-tailored specialisation effectively enables a simple and interpretable similarity-based classification based on nearest neighbours (NNs) in the specialised encoder space (see Q2 and Figure~\ref{fig:tsne-ocl}). 

The two-stage \convfit transformation offers new insights and contributions to representation learning for conversational tasks. Unlike prior work which conducted large-scale conversational pretraining from scratch using large datasets, we demonstrate that full pretraining is not needed to obtain universal conversational encoders. By leveraging the general semantic knowledge already stored in pretrained LMs, we can expose (i.e., 'rewire') that knowledge \cite{Vulic:2021acl,Gao:2021arxiv,Liu:2021emnlp} via much cheaper and quicker adaptive fine-tuning on a tiny fraction of the full Reddit data (e.g., even using $<0.01\%$ of the Reddit corpus). Further, the task-oriented S2 \convfit-ing transforms pretrained LMs into task-specialised sentence encoders. Our results with similarity-based classification, targeting the crucial conversational NLU task of intent detection (ID), reach state-of-the-art (SotA) across all standard ID datasets, with particular gains in the most challenging, few-shot setups. Importantly, we show that the gradual application of S1 and then S2 yields a synergistic effect, that is, it attains the highest ID results across the board.

Finally, \convfit is highly versatile: it can be used with a range of pretrained LMs and on a spectrum of text classification problems; it also allows for the simple usage of diverse fine-tuning objectives in both Stage 1 and Stage 2, beyond the ones proposed and evaluated in this work. 







%% file: 02-methodology.tex



\noindent \textbf{Preliminaries.} For any input text $t$, we obtain its encoding $\mathbf{t}={enc}(t)$, where $enc$ is a sentence encoder at any \convfit stage (i.e., before any fine-tuning, after S1, or after S2), or any other sentence encoder. The text $t$ is tokenized into subwords \cite{Schuster2012} relying on each encoder's dedicated tokeniser. The final encoding $\textbf{t}$ is created via a \textit{pooling} operation such as (a) using the [CLS] token, (b) or mean-pooling the output subword vectors. Following prior work \cite{Reimers:2019emnlp}, we always use mean-pooling.

\subsection{Stage 1: Adaptive Fine-Tuning}
\label{ss:stage1}


As in prior work on conversational pretraining \cite{Henderson:2019acl,Henderson:2020convert,Humeau:2019arxiv}, Stage 1 relies on the response ranking task with Reddit data and dual-encoder architectures, which model the interaction between Reddit \textit{(context, response)} $(c,r)$ pairs.\footnote{In each $(c,r)$ pair, $r$ is the \textit{response} that immediately follows the preceding \textit{context} sentence in a Reddit thread; see \cite{Henderson:2019arxiv}. The intuition is that sentences which elicit similar responses should obtain similar sentence encodings \cite{Yang:2018repl}.}  However, unlike prior work, instead of pretraining from scratch we fine-tune an LM-pretrained encoder, which yields a much quicker conversational encoder specialisation, and does not require massive amounts of data. 

Response ranking is formulated as the standard multiple negatives ranking loss (\textsc{mneg}): for each positive $(c_i,r_i)$ pair (i.e., the pair observed in the Reddit fine-tuning data), the aim is to rank the correct response $r$ for the input $c$ over a set of randomly sampled responses $r_j, j \neq i$ from other Reddit pairs. The similarity between $c$-s and $r$-s is quantified via the similarity function $S$ operating on their encodings $S(\mathbf{c}, \mathbf{r})$. Following prior work, we use the scaled cosine similarity: $S(\mathbf{c}, \mathbf{r}) = D\cdot cos(\mathbf{c}, \mathbf{r})$, where $D$ is the scaling constant. Stage 1 fine-tuning with \mneg then proceeds in batches of $B$ positive Reddit pairs $(c_i,r_i), \ldots, (c_B,r_B)$; the \mneg loss for a single batch is computed as:

\vspace{-2mm}
{\small
\begin{align}
\mathcal{L} = -\sum_{i=1}^B S(\mathbf{c_i},\mathbf{r_i}) + \sum_{i=1}^B \log \sum_{j=1,j\neq i}^{B} e^{S(\mathbf{c_i},\mathbf{r_j})}
\label{eq:mneg}
\end{align}
}%
\noindent Effectively, for each batch Eq.~\eqref{eq:mneg} maximises the similarity score of positive context-response pairs $(c_i, r_i)$, while it minimises the score of $B-1$ random pairs. The negative examples are all pairings of $c_i$ with $r_j$-s in the current batch, where such $(c_i, r_j)$ pairs do not occur in the Reddit data.\footnote{We also experimented with another SotA loss function, the triplet-based multi-similarity loss \cite{Wang:2019cvpr,Liu:2020arxiv}, without any substantial performance differences.} 

The output of Stage 1 is the sentence encoder $enc_{S1}$ which can be used 'as is' similarly to standard sentence encoders \cite{Henderson:2020convert,Casanueva:2020ws,Feng:2020labse}: a standard ID approach stacks a Multi-Layer Perceptron (MLP) classifier on top of the fixed sentence vectors $\mathbf{t}$, and fine-tunes only the MLP parameters \cite{Casanueva:2020ws,Gerz:2021arxiv}. However, the output of S1 can also be further fed as the input encoding for \convfit's Stage 2 (Figure~\ref{fig:overview}).

\subsection{Stage 2: Task-Based Sentence Encoders}
\label{ss:stage2}
Stage 2 fine-tuning is inspired by metric-based meta-learning \cite{Vinyals:2016nips,Musgrave:2020eccv} and exemplar-based (also termed prototype-based) learning \cite{Snell:2017nips,Sung:2018cvpr,Zhang:2020emnlp}, which is especially suited for few-shot scenarios. We assume the existence of $N_a$ annotated in-task examples $\{(x_1,y_1),\ldots,(x_{N_a},y_{N_a})\}$: e.g., $x$-s are text sentences with $y$-s being their intent labels/classes; let us assume that there are $N_c$ classes $\{C_1,\ldots,C_{N_c}\}$ in total. The aim is to fine-tune the input sentence encoder in such a way to encode all sentences associated with each particular class into coherent clusters, clearly separated from all other class-related (also coherent) clusters (see Figure~\ref{fig:tsne-ocl}).\footnote{In other words, the encoder should learn to encode each utterance into one of such semantically well-defined clusters.} 

\vspace{1.3mm}
\noindent \textbf{Positive and Negative Pairs.}
We leverage the class labels only implicitly (see Figure~\ref{fig:overview}), which allows us to \textit{treat intent detection as a sentence similarity task}. \convfit S2 operates with two sets of pairs: \textbf{1)} $PP$ is the set of positive pairs $(x_i, x_j)$, where $x_i$ and $x_j$ are text instances associated with the same class $C_i$; \textbf{2)} $NP$ contains negative pairs $(x_i, x_j)$ where $x_i$ and $x_j$ are associated with two different classes $C_i$ and $C_j$. We construct the set $NP$ in a balanced way: for each positive pair $(x_i, x_j) \in PP$, we add $2\times n$ negative pairs into $NP$, where $n$ is a tunable hyper-parameter; $n$ pairs $(x_i, x_{i,n'}), n'=1,\ldots,n$, are constructed by randomly sampling utterances $x_{i,n'}$ which do not share the class with $x_i$, and we also sample $n$ negatives $(x_{j,n'}, x_j)$ in a similar vein. We now present three different loss functions that fine-tune the input encoders towards task-specialised sentence similarity relying on the sets $PP$ and $NP$. For all three S2 loss functions, we add a \textit{down-projection} $d_{o}$-dim layer with non-linearity (\textit{Tanh} used) after pooling, see Figure~\ref{fig:overview}.\footnote{A variant with down-projection yielded slightly higher scores than the one without it in our preliminary experiments.} 

\vspace{1.2mm}
\noindent \textbf{\softmax (\smax) Loss.} 
Following prior work \cite{Reimers:2019emnlp}, for each input sentence pair $(x_i, x_j)$, we concatenate their $d_{o}$-dimensional encodings $\mathbf{x_i}$ and $\mathbf{x_j}$ (obtained after passing them through the input encoder, pooling, and down-projection) with their element-wise difference $|\mathbf{x_i} -  \mathbf{x_j}|$. The objective is as follows: $\mathcal{L}_{\smax} = \text{softmax}\big(W(\mathbf{x_i} \oplus \mathbf{x_j} \oplus |\mathbf{x_i}-\mathbf{x_j}|)\big)$,
 where $\oplus$ denotes concatenation, and $W \in \mathbb{R}^{3d_{o} \times 2}$ is a trainable weight matrix of the softmax classifier, where $2$ is the number of classification classes: the model must simply discern between positive pairs (from $PP$) and negative pairs from $NP$. The classifiers are optimised via standard cross-entropy. 

\vspace{1.2mm}
\noindent \textbf{Cosine (\cosine) Loss.} 
The idea is to minimise the following distance, formulated as standard mean-squared error: $||\delta_{l} - cos(\mathbf{x_i},\mathbf{x_j})||_2$, where $cos$ denotes cosine similarity, and $\delta_{l}$ is a hyper-parameter which specifies the 'ideal' (dis)similarity margin in the specialised encoder space. Here, we rely on the default parameters from \newcite{Reimers:2019emnlp} without any tuning: $\delta_{l}=0.8$ iff $(x_i,x_j) \in PP$, and $\delta_{l}=0.3$ iff $(x_i, x_j) \in NP$.

\vspace{1.2mm}
\noindent \textbf{Online Contrastive Learning (\nce) Loss} follows the formulation from \newcite{Hadsell:2006cvpr}:

\vspace{-2mm}
{\footnotesize
\begin{align}
\mathcal{L}_{\nce} & =  \mathds{1} \cdot (dcos(\mathbf{x_i}, \mathbf{x_j}))^{2} \notag \\
&+ (1-\mathds{1})\cdot \Big(ReLU(\delta_{m}-dcos(\mathbf{x_i}, \mathbf{x_j}))\Big)^{2}
\end{align}}%
where $\mathds{1}$ is the indicator function which returns 1 iff $(\mathbf{x_i}, \mathbf{x_j}) \in PP$, and 0 iff $(\mathbf{x_i}, \mathbf{x_j}) \in NP$; $dcos = 1 - cos$ is the cosine distance, and $\delta_{m}$ is the distance margin, set to the default value of 0.5 \cite{Reimers:2019emnlp} in all our experiments. The loss 'attracts' similar items closer together in the specialised space, while 'repelling' dissimilar items \cite{Mrksic:17}.\footnote{We use the \textit{online} version of the loss that updates the loss focusing on hard negative pairs (i.e., negatives that are close by cosine in the current semantic space) and hard positives which are far apart in the current space. This typically results in quicker convergence and slightly better performance.}

\vspace{1.3mm}
\noindent \textbf{Similarity-Based Inference.} Intent detection in the specialised encoder space $enc_{S2}$ is then performed via similarity-based classification \cite{Zhang:2020emnlp} after Stage 2.\footnote{The benefits of similarity-based classification were recently validated also in other NLP tasks such as cross-lingual abusive content detection \cite{Sarwar:2021arxiv}, language modeling \cite{Khandelwal:2020iclr,Guu:2020icml}, and question answering \cite{Kassner:2020emnlp}, among others.} Assuming the simplest case of $k=1$ nearest neighbours (NN) classification, we select the intent class for an unseen example $u$ as: $I_c\big(\arg \max_{\mathbf{t} \in Pool} cos(\mathbf{t}, \mathbf{u})\big)$. Here, $\mathbf{t}=enc_{S2}(t)$ refers to the sentence encoding of each example $t \in Pool$ (which is typically the pool of examples from the ID training set), and the $I_c$ function returns the intent class of any $t \in Pool$.

\vspace{1.3mm}
\noindent \textbf{Why Intent Detection as a Sentence Similarity Task?} 
 We can take the analogy of `intent' being a latent semantic class where sentences associated with the intent are diverse surface instances of the class (i.e., language realisations of the underlying concept/intent). This means that finding the most similar labelled instances for the given unlabelled input instance/sentence can directly inform us about the underlying semantic class/intent. 
 



%% file: 03-exp.tex
\noindent \textbf{Input LMs.} We experiment with several popular Transformer-based \cite{Vaswani:2017nips} LMs as input (see Figure~\ref{fig:overview}), aiming to validate the robustness of \convfit, as well as to analyse the impact of LM pretraining on the final task performance: (i) BERT \cite{Devlin:2018arxiv} (labeled \bert henceforth); (ii) RoBERTa (\rob), as an improved variant of BERT, LM-pretrained with more data \cite{Liu:2019roberta}; (iii) DistilRoBERTa (\drob), a distilled more compact version of RoBERTa, LM-pretrained with around 4 times fewer data than the teacher RoBERTa model \cite{Sanh:2019arxiv}. The cased \textsc{Base} variants are used for all input LMs: 768-dimensional Transformer layers with 12 (\bert, \rob) or 6 (\drob) attention layers. In addition, to isolate the effects of LM-pretraining and \convfit-ing from the mere ``parameter capacity'', we also experiment with a \bert/\rob architecture with \textsc{rand}omly initialised parameters using the Xavier initialisation \cite{Glorot:2010aistats}.

Unless noted otherwise, \convfit Stage 1 always proceeds with a sample comprising 2\% of the full Reddit corpus from \newcite{Henderson:2019arxiv}.\footnote{The full corpus contains 700M+ \textit{(context, response)} pairs.}

\begin{table}[t]
\fontsize{7.9}{10}\selectfont
\def\arraystretch{0.66}
\centering
\begin{tabularx}{\linewidth}{l XXX}
\toprule
\textbf{Dataset} & \textbf{Intents} & \textbf{Examples} & \textbf{Domains}\\
\midrule
{\textsc{banking77}} & 77 & 13,083 & 1 (banking)  \\
{\textsc{clinc150}} & 150 & 23,700 & 10 \\
{\textsc{hwu64}} & 64 & 25,716 & 21 \\

\bottomrule
\end{tabularx}
\vspace{-1.5mm}
\caption{Intent detection datasets: key statistics.}
\label{tab:intent-data}
\vspace{-2.5mm}
\end{table}

\vspace{1.3mm}
\noindent \textbf{Intent Detection Datasets.} As discussed in \S\ref{s:methodology}, the main evaluation task is intent detection (ID), with a particular focus on low-data (i.e., few-shot) scenarios. Our Stage 2 fine-tuning and the final task evaluation are based on three standard ID datasets in English, also available as part of the recently published DialoGLUE benchmark \cite{Mehri:2020dialoglue}: \banking \cite{Casanueva:2020ws}, \hwu \cite{Liu:2019iwsds}, and \clinc \cite{Larson:2019emnlp}.\footnote{The datasets provide a range of diverse ID setups, covering fine-grained ID within a single domain (e.g., \banking), as well as coarser-grained ID spanning several well-defined domains (e.g., news, calendar, alarm, restaurant booking in \hwu or in \clinc). They provide a more challenging setup (and are also better aligned with the actual ID setups typically met in production) than some other well-known ID datasets such as SNIPS \cite{Coucke:18}.} The key statistics of all three datasets are provided in Table~\ref{tab:intent-data}; for further details, we refer the reader to the original work and also to \cite{Mehri:2020dialoglue}. 

\vspace{1.3mm}
\noindent \textbf{Few-Shot and Full Data Setups.}
Prior work has recognised the importance of building intent detectors in low-data regimes \cite{Casanueva:2020ws,Mehri:2021naacl}. Therefore, following this initiative, we evaluate the models in two \textbf{N-shot} scenarios, where we assume that only $N=10$ or $N=30$ annotated examples per intent are available for training the MLP classifier or for S2 fine-tuning; Figure~\ref{fig:overview}.\footnote{We use the same fixed few-shot and test sets for each intent detection dataset as released by \newcite{Mehri:2020dialoglue}.} The models are also evaluated in the \textbf{Full} setup, where all annotated training examples per intent are used. Note that we always report the scores on the same test set for each setup. For the few-shot scenarios, we report the scores as averages over 3 independent experimental runs.

\vspace{1.3mm}
\noindent \textbf{Hyperparameters and Optimisation.}
\convfit is implemented via the \textit{sentence-transformers (sbert)} repository \cite{Reimers:2019emnlp}, which is in turn built on top of the HuggingFace repository \cite{Wolf:2019hf}. Similar to \newcite{Casanueva:2020ws}, we do not rely on any development data, and follow the general suggestions from prior work \cite{Reimers:2019emnlp,Casanueva:2020ws} for the hyperparameter setup, which is adopted across all intent ID datasets.\footnote{For all MLP intent classifiers, this implies relying on the empirically validated and stable setup from prior work \cite{Casanueva:2020ws}: the best results are achieved with a 2-layer fully-connected MLP ($768$-dim hidden layers), trained via SGD with the high learning rate (0.5) and linear decay, and very aggressive dropout rates (0.75); training lasts for 500 epochs; batch size is 32. This setup achieved strong results in our preliminary experiments as well, and is thus adopted here.} For S1 with \mneg, we always train for 2 epochs in batches of 256 with default hparams from \textit{sbert}.\footnote{256 is the maximum batch size with \textsc{Base} BERT and RoBERTa which allows us to run Stage 1 fine-tuning on a single 12GiB GTX GPU.} 

In Stage 2, with all three evaluated objective functions the batch size is 32, the maximum sequence length is 48, the output layer's dimensionality is set to $d_{o}=512$. Unless stated otherwise, we always fine-tune for 10, 5, and 2 epochs for the 10-shot, 30-shot, and Full setups, respectively. For the \cosine and \nce variants, unless noted otherwise, we report the results with $n=3$ negative examples per each positive in 10-shot and 30-shot setups, and with $n=1$ (for computational tractability) in the Full setup. An analysis of the impact of $n$ on the final ID performance is presented later in \S\ref{s:results}.


Following the suggested settings of \newcite{Reimers:2019emnlp,Vulic:2020emnlp}, in both \convfit stages we use the AdamW optimiser \cite{Loschilov:2018iclr}; the learning rate is $2e-5$ with the warmup rate of $0.1$ and linear decay afterwards, and the weight decay rate is set to $0.01$. 


\vspace{1.3mm}
\noindent \textbf{Similarity-Based Classification.}
The intent class is chosen according to the $k=1$ NNs, based on the cosine distance in the fine-tuned space.\footnote{Very similar results are observed with $k=3$ and $k=5$.} Importantly, in few-shot setups we use \textit{only} the few-shot data as the NN pool for classification.

\begin{table*}[t]
\def\arraystretch{0.7}
\centering
{\scriptsize
\begin{tabularx}{\linewidth}{l XXX XXX XXX}
\toprule
  {} & \multicolumn{3}{c}{\bf \textsc{banking77}} & \multicolumn{3}{c}{\bf \textsc{clinc150}} & \multicolumn{3}{c}{\bf \textsc{hwu64}} \\
  \cmidrule(lr){2-4} \cmidrule(lr){5-7} \cmidrule(lr){8-10}
\textbf{Model Variant} & \textbf{10}  & \textbf{30} & \textbf{Full} & \textbf{10}  & \textbf{30} & \textbf{Full} & \textbf{10}  & \textbf{30} & \textbf{Full} \\
\cmidrule(lr){2-10}
 \cmidrule(lr){1-10}
 \rowcolor{Gray}
 {} & \multicolumn{9}{c}{\bf Similarity-Based Classification} \\
  \cmidrule(lr){2-10}
 \textsc{rob+s1+s2-cos} & \underline{86.48} & \underline{91.33} & {\bf 94.35} & \underline{92.87} & \underline{95.91} & \underline{97.20} & {85.06} & {\bf 90.46} & {\bf 92.98} \\
 \textsc{bert+s1+s2-cos} & {84.32} & {90.91} & {93.91} & {91.80} & {95.58} & {96.56} & \underline{85.13} & {89.41} & {91.93} \\
 \textsc{drob+s1+s2-cos} & {85.13} & {90.75} & {94.06} & {91.64} & {95.48} & {97.00} & {83.64} & {89.68} & \underline{92.94} \\
 \cmidrule(lr){2-4} \cmidrule(lr){5-7} \cmidrule(lr){8-10}
 \textsc{rob+s1+s2-ocl} & {\bf 87.38} & {\bf 91.36} & {94.16} & {\bf 92.89} & {\bf 96.42} & {\bf 97.34} & {\bf 85.32} & \underline{90.06} & {92.42} \\
 \textsc{bert+s1+s2-ocl} & {85.97} & {90.65} & {93.77} & {91.53} & {95.53} & {96.82} & {85.04} & {89.41} & {92.21} \\
 \textsc{drob+s1+s2-ocl} & {86.04} & {90.78} & {93.89} & {91.98} & {95.60} & {97.04} & {83.64} & {89.50} & {92.84} \\
 \cmidrule(lr){2-4} \cmidrule(lr){5-7} \cmidrule(lr){8-10}
 \textsc{rob+s2-cos} & {84.96} & {90.81} & \underline{94.19} & {91.56} & {95.64} & {96.78} & {84.52} & {89.87} & {92.19} \\
 \textsc{bert+s2-cos} & {81.27} & {90.32} & {93.73} & {89.58} & {95.08} & {96.54} & {82.90} & {89.12} & {91.78} \\
 \textsc{drob+s2-cos} & {83.28} & {90.58} & {93.91} & {89.47} & {95.32} & {86.78} & {82.43} & {89.41} & {92.10} \\
 \cmidrule(lr){2-4} \cmidrule(lr){5-7} \cmidrule(lr){8-10}
 \textsc{rob+s2-ocl} & {85.78} & {90.98} & {93.77} & {92.64} & {95.40} & {96.87} & {84.76} & {89.31} & {92.01} \\
 \textsc{bert+s2-ocl} & {82.28} & {89.77} & {93.54} & {90.71} & {95.07} & {96.62} & {83.09} & {88.94} & {92.57} \\
 \textsc{drob+s2-ocl} & {82.60} & {90.65} & {93.38} & {90.78} & {95.02} & {96.69} & {81.69} & {88.75} & {92.38} \\
 \cmidrule(lr){2-10}
 \rowcolor{Gray}
 {} & \multicolumn{9}{c}{\bf Baselines: MLP Classification} \\
\cmidrule(lr){2-10}
\textsc{rob+s1} & {83.08} & {90.16} & {93.38} & {90.98} & {94.12} & {96.42} & {81.13} & {87.73} & {91.44} \\
\textsc{bert+s1} & {82.69} & {89.82} & {93.67} & {89.88} & {94.07} & {96.33} & {82.25} & {88.01} & {91.12} \\
\cmidrule(lr){2-4} \cmidrule(lr){5-7} \cmidrule(lr){8-10}
\textsc{ConveRT$^*$} & {83.32} & {89.37} & {93.01} & {92.62} & {95.78} & {97.16} & {82.65} & {87.88} & {91.24} \\
\textsc{USE$^*$} & {84.23} & {89.74} & {92.81} & {90.85} & {93.98} & {95.06} & {83.75} & {89.03} & {91.25} \\
\textsc{USE} (ours) & {82.95} & {89.09} & {92.81} & {90.27} & {93.54} & {94.91} & {82.71} & {88.20} & {91.64} \\
\textsc{LaBSE} & {81.69} & {88.96} & {92.60} & {90.89} & {93.41} & {95.12} & {81.60} & {86.15} & {90.99} \\

\bottomrule
\end{tabularx}
}%
\vspace{-1.5mm}
\caption{Accuracy scores ($\times$100\%) on the three ID data sets with varying number of training examples (\textbf{10} examples per intent; \textbf{30} examples per intent; \textbf{Full} training data). $n=3$ negatives are used in Stage 2 for 10-shot and 30-shot setups, $n=1$ for the Full setup (see \S\ref{s:experimental}). The peak scores per column are in bold, the second best is underlined. *The scores were taken directly from prior work, and computed on different 10/30-shot samples (and are thus not directly comparable, \citealt{Zhao:2020arxiv}). For clarity, we show only a subset of (arguably most informative) model variants; the complete table with additional evaluated variants is available in the Appendix.}
\label{tab:main}
\vspace{-1.5mm}
\end{table*}

\subsection{Model Variants and Baselines}
\label{ss:variants}
We experiment with a range of model variants enabled by the \convfit framework (see Figure~\ref{fig:overview}), and compare their performance in the ID task against an array of cutting-edge universal and conversational sentence encoders. All the models in evaluation are summarised here for clarity.

\vspace{1mm}
\noindent {\bf \textsc{lm+s1+s2-loss}.} Sentence encoders after running the full \convfit pipeline, where intent detection is based on similarity-based NN classification. \textsc{lm} in the label of this variant denotes the input LM, and \textsc{loss} is the loss function used in Stage 2 (i.e., \textsc{smax}, \textsc{cos}, or \textsc{ocl}).

\vspace{1mm}
\noindent {\bf \textsc{lm+s2-loss}.} Sentence encoders optimised only via Stage 2 \convfit, skipping Stage 1 (see Figure~\ref{fig:overview}); similarity-based intent detection.

\vspace{1mm}
\noindent {\bf \textsc{lm+s1}.} The input LM is converted into a (general-purpose) conversational encoder via Stage 1 \convfit-ing; intent detection is performed via standard feature-based MLP classification on top of the sentence encodings as in prior work. 

\vspace{1mm}
\noindent {\bf SotA Sentence Encoders.} We evaluate three widely used state-of-the-art sentence encoders in the standard feature-based MLP classification approach to intent detection:\footnote{For more technical details regarding each sentence encoder, we refer the reader to the original work.} (i) \textit{ConveRT} \cite{Henderson:2020convert} is a dual sentence encoder pretrained with the conversational response selection task \cite{Henderson:2019acl} on the full Reddit data \cite{AlRfou:2016arxiv,Henderson:2019arxiv}; (ii) multilingual Universal Sentence Encoder (\textit{mUSE}) \cite{yang2020mUSE} is a multilingual and better-performing version of the USE model for English \cite{Cer:2018arxiv}, which again relies on a standard dual-encoder framework \cite{Henderson:2019acl,Humeau:2019arxiv} and is pretrained on massive amounts of data; (iii) Language-agnostic BERT Sentence Embedding (\textit{LaBSE}) \cite{Feng:2020labse} adapts pretrained multilingual BERT (mBERT) \cite{Devlin:2018arxiv} into a sentence encoder using a dual-encoder framework \cite{yang2019ijcai} with larger embedding capacity (i.e., it provides a shared multilingual vocabulary spanning 500k subwords).\footnote{LaBSE is the current SotA encoder across a wide array of languages \cite{Feng:2020labse,Litschko:2021ecir,Gerz:2021arxiv}. Besides dual-encoder training, LaBSE leverages standard self-supervised objectives used in pretraining of mBERT and XLM: masked and translation language modeling \cite{Conneau:2019nips}; see the original work.}


%% file: 04-results.tex

The main results are summarised in Table~\ref{tab:main}, and further results and analyses are available in \S\ref{ss:further}, with additional results in the Appendix.\footnote{For brevity, in the main paper we report the results with the two better-performing S2 losses: \cosine and \nce.} These results offer multiple axes of comparison, succinctly discussed in what follows.

\vspace{1.3mm}
\noindent \textbf{MLP versus Similarity-Based ID.}
First, we note that \convfit-ed LMs achieve peak ID scores across all three ID datasets, and in all data setups, with \textsc{rob+s1+s2-ocl} being the highest-performing model variant overall. Running Stage 1 does transform input LMs into effective (universal) conversational encoders already: for MLP-based ID, we observe competitive or even improved performance (cf., the results on \banking and \hwu as two more challenging evaluation sets) with the \textsc{rob+s1} and \textsc{bert+s1} variants against current state-of-the-art (conversational) sentence encoders such as ConveRT, USE, and LaBSE. 

Importantly, the results after Stage 2 `unanimously' suggest the effectiveness of treating ID as a semantic similarity task, and additional task-specific specialisation of the sentence encoders with in-task data. Put simply, it seems more effective to use the in-task training data to `task-specialise' the sentence encoder space than to learn a standard (MLP) classifier, which directly maps from the feature space to intent labels \cite{Sarwar:2021arxiv}. The gains are especially pronounced in few-shot setups (e.g., see 10-shot \banking). 

We speculate that dual-encoder contrastive learning surpasses MLP-based approaches especially in few-data scenarios because it learns from finer-grained and more abundant information in such low-data scenarios: i.e., we learn to contrast between pairs of instances rather than simply learning an MLP-based mapping from an instance to its underlying class intent/class. This formulation can also capture some subtle cross-instance (dis)similarities which cannot be captured by MLP. Extending beyond pure absolute performance, decisions based on $k$-NN similarity-based ID in the specialised space are also easy to interpret \cite{Simard:1992,Wallace:2018ws}.

\vspace{1.3mm}
\noindent \textbf{Stage 1 + Stage 2?}
The scores in Table~\ref{tab:main} indicate that Stage 2 alone already transforms pretrained LMs into very strong task-specialised sentence encoders. However, a more careful comparison of \textsc{lm+s1+s2-loss} versus \textsc{lm+s2-loss} variants reveals that Stage 1 fine-tuning is universally useful (regardless of the chosen loss function in S2), and yields ID performance gains. In other words, the coarser-grained adaptive fine-tuning already exposes some conversational knowledge from the pretrained LMs, and such knowledge does have substantial impact on task-specialised S2 tuning. In sum, this finding is line with prior work in other domains and NLP tasks \cite{Gururangan:2020acl,Glavas:2020coling,Ruder:2021blog}: both domain-adaptive (our S1) and task-adaptive  additional tuning (our S2) of general-purpose LMs have a synergistic positive impact on the final task performance.

\captionsetup[subfigure]{oneside,margin={-1.5cm,-2.5cm},skip=-4pt}
\begin{figure}[!t]
    \centering
    \includegraphics[width=0.86\columnwidth]{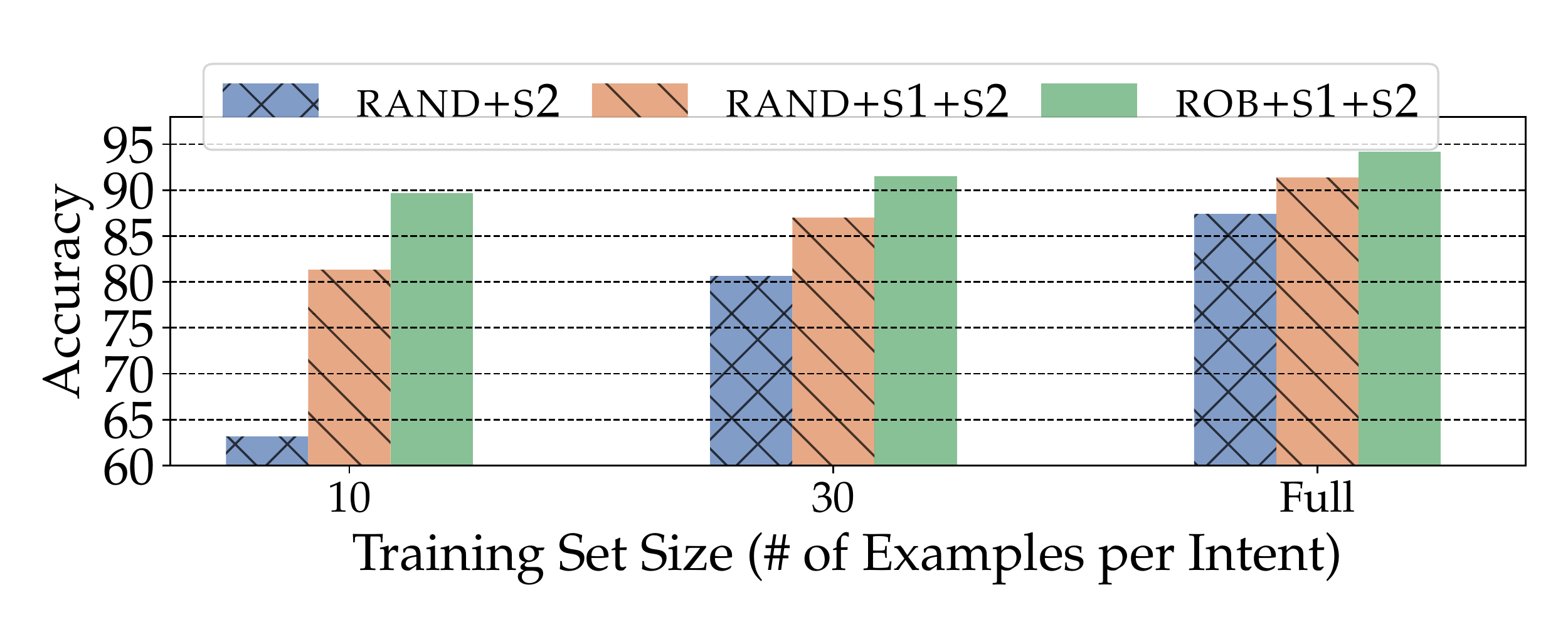}
    \vspace{-2.5mm}
    \caption{A comparison of a randomly initialised RoBERTa (\textsc{rand}) against LM-pretrained RoBERTa after S2 \convfit-ing with \nce; \banking.}
    \vspace{-4mm}
    \label{fig:rand-main}
\end{figure}
\captionsetup[subfigure]{oneside,margin={-0.5cm,0.5cm},skip=-2pt}
\begin{figure*}[t!]
    \centering
    \begin{subfigure}[!ht]{0.414\linewidth}
        \centering
        \includegraphics[width=0.98\linewidth]{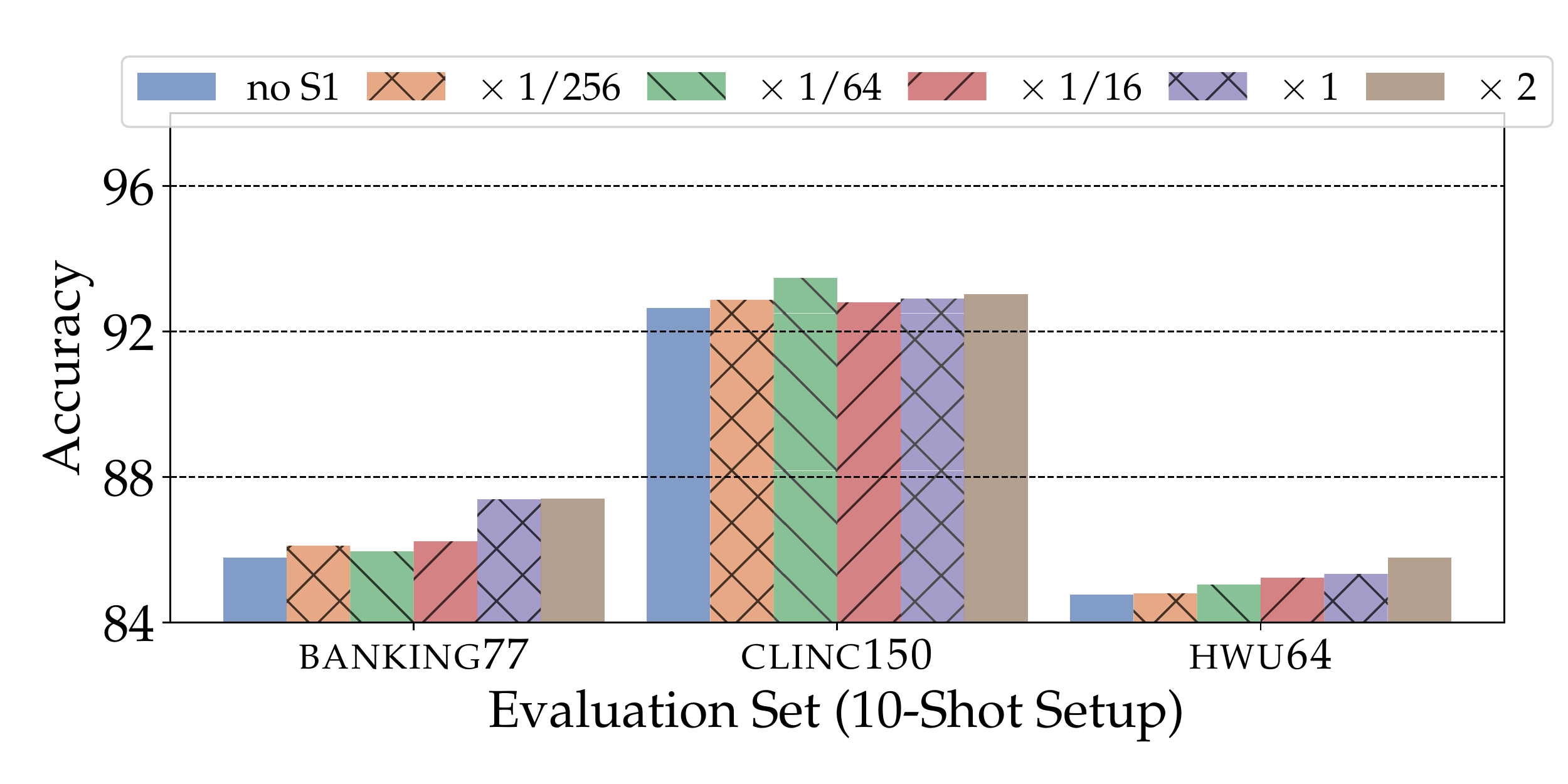}
        \caption{10-shot (\textsc{rob+s1+s2-ocl})}
        \label{fig:sizes-ocl-10}
    \end{subfigure}
    \begin{subfigure}[!ht]{0.414\textwidth}
        \centering
        \includegraphics[width=0.98\linewidth]{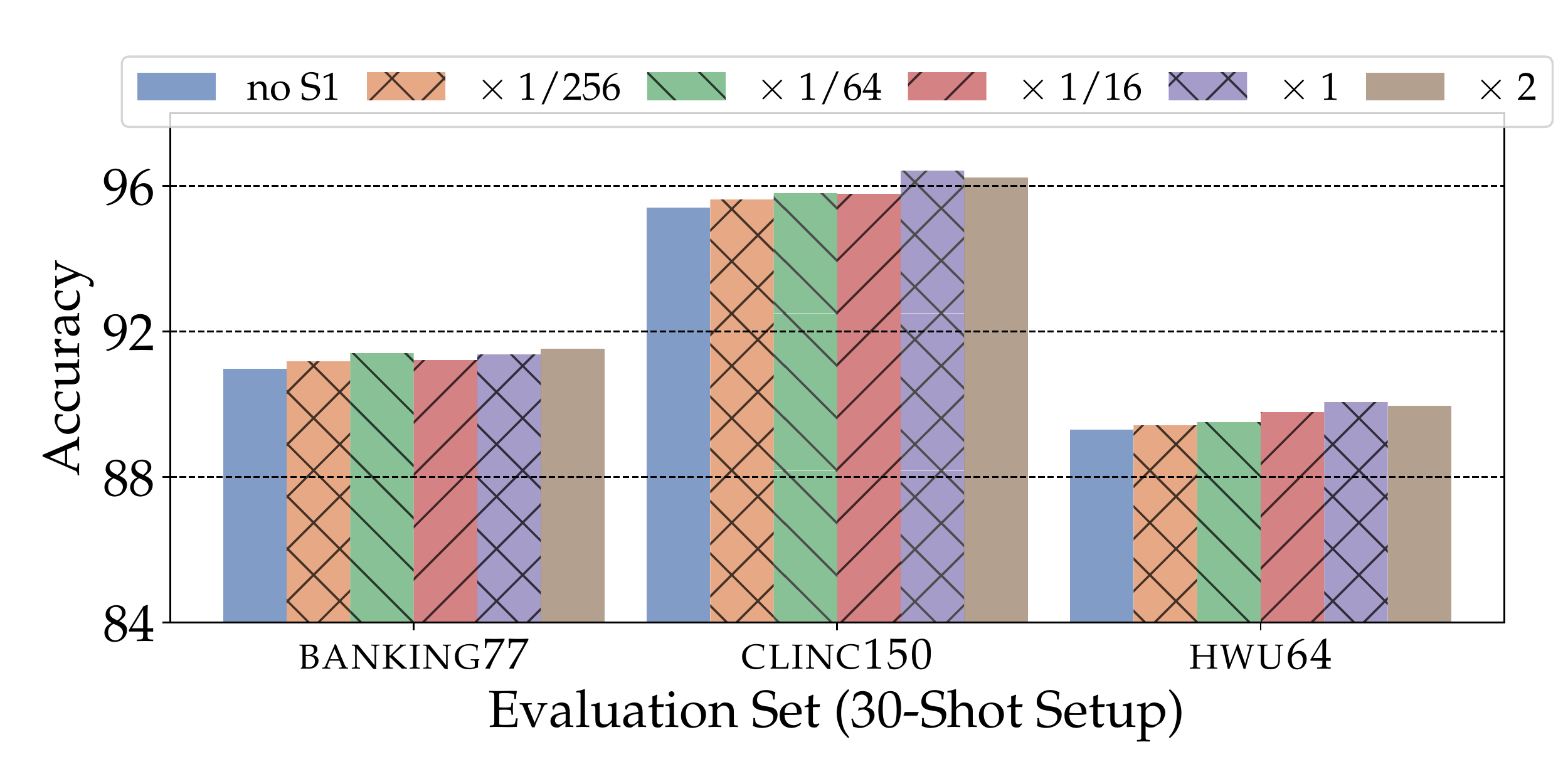}
        \caption{30-shot (\textsc{rob+s1+s2-ocl})}
        \label{fig:sizes-ocl-30}
    \end{subfigure}
    \vspace{-1.5mm}
    \caption{Varying the amount of Reddit data for Stage 1 \convfit; $\times 1$ refers to the Reddit size used in all our other Stage 1 fine-tuning experiments ($\approx$2\% of the full Reddit corpus from \newcite{Henderson:2019arxiv}), while other Reddit data sizes are relative to this corpus size (e.g., $\times 1/32$ means that we use $2\%/32\approx0.0625\%$ of the full Reddit corpus). Similar plots (with similar findings) using the \cosine loss in Stage 2 are available in the Appendix.}
    \vspace{-2mm}
\label{fig:sizes-ocl}
\end{figure*}
\begin{figure}[!t]
    \centering
    \begin{subfigure}[!ht]{0.489\linewidth}
        \centering
        \includegraphics[width=0.999\linewidth]{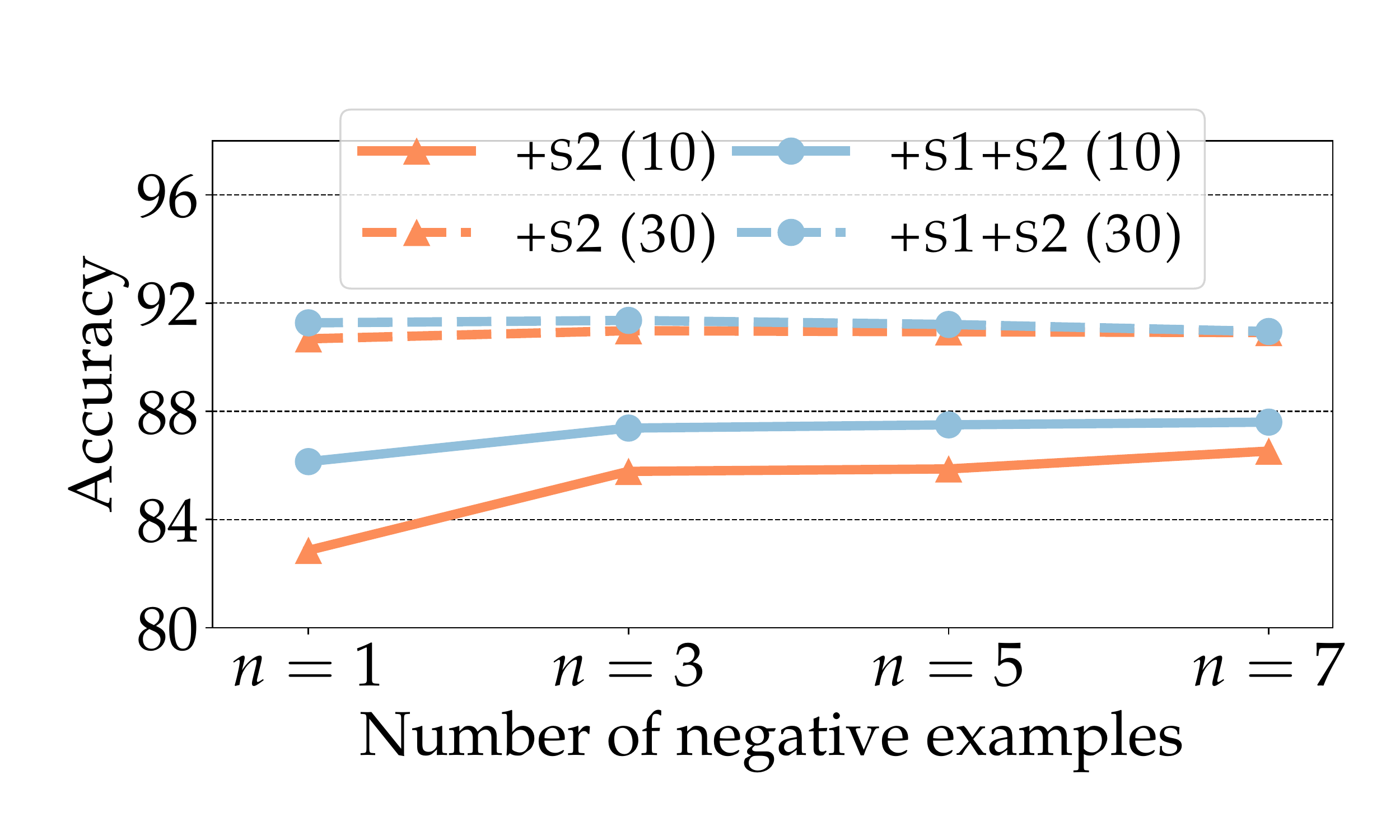}
        \caption{\banking}
        \label{fig:neg-ocl-banking}
    \end{subfigure}
    \begin{subfigure}[!ht]{0.489\linewidth}
        \centering
        \includegraphics[width=0.999\linewidth]{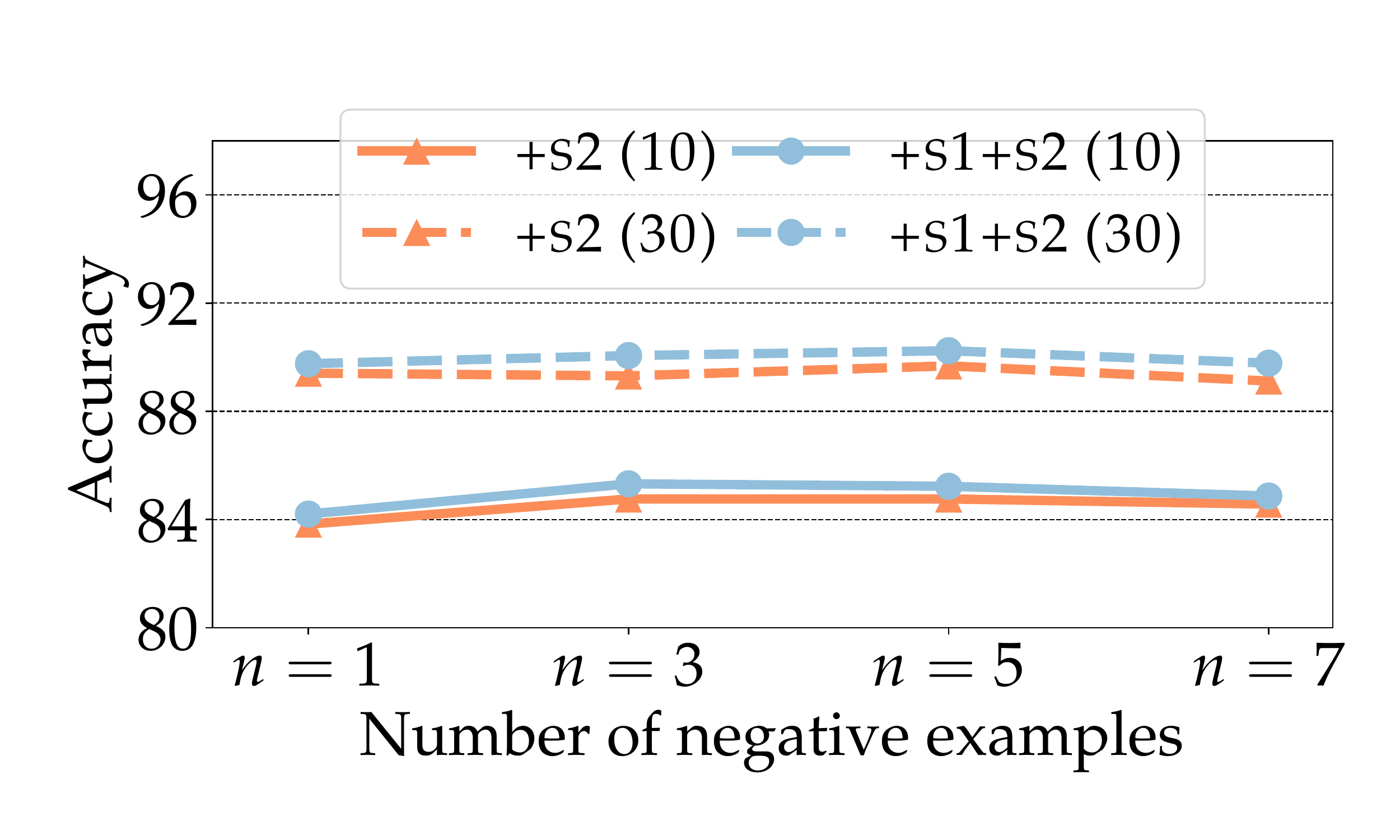}
        \caption{\hwu}
        \label{fig:neg-ocl-hwu}
    \end{subfigure}
    \vspace{-2.3mm}
    \caption{Impact of the number of negative examples $n$ in 10-shot and 30-shot setups. The \convfit variants are \textsc{rob+s2-ocl} and \textsc{rob+s1+s2-ocl} (labelled +S1 and +S1+S2 in the figures, respectively).}
    \vspace{-2.5mm}
\label{fig:neg-ocl}
\end{figure}

The impact of the gradual two-stage sentence encoder transformation is also clearly visible from the t-SNE visualisation in Figure~\ref{fig:tsne-rob}. Besides this, a standard quantitative measure of cluster coherence, the Silhouette coefficient $\sigma$ \cite{silhouette} also points in the same direction: $\sigma=0.067$ for the test examples and model variant from Figure~\ref{fig:tsne-orig}, $\sigma=0.188$ (Figure~\ref{fig:tsne-r15m}), and $\sigma=0.698$ (Figure~\ref{fig:tsne-ocl}).\footnote{Higher $\sigma$ scores are desirable as they imply more coherent and compact clusters, and a stronger inter-cluster separation.} \footnote{Stage 2 tuning with more in-task data also naturally yields a better separation of examples into coherent clusters , which then naturally improves NN-based classification. For instance, running the \textsc{rob+s1+s2-ocl} ($n=3$) variant in 10-shot, 30-shot, and Full data setups yields the respective $\sigma$ scores for the same set of test examples from Figure~\ref{fig:tsne-rob}: $\sigma_{10}=0.378$, $\sigma_{30}=0.548$, $\sigma_{Full}=0.698$, validating the intuition.}

\vspace{1.3mm}
\noindent \textbf{Impact of Input LMs.} While the results suggest that the \convfit framework is applicable and effective with any pretrained LM, the choice of the input LM naturally impacts the absolute ID performance. As expected, the \convfit variants with RoBERTa achieve the highest scores across the board. A comparison between \drob and \bert reveals that the pretraining data size and regime seem to play a more critical role than the parameter capacity: the more compact \drob LM is competitive with or even outscores \bert-based variants.\footnote{Given the versatility of \convfit, in future work we plan to extend the experiments to other pretrained LMs such as ELECTRA \cite{Clark:2020iclr} and T5 \cite{Raffel:2020t5}.}

\vspace{1.3mm}
\noindent \textbf{Importance of LM Pretraining} is illustrated by Figure~\ref{fig:rand-main}. The trend is quite straightforward: semantic knowledge acquired by LM-pretraining is particularly important in the fewest-shot (i.e., 10-shot) setups, and the gap gets reduced with more in-task data available for S2 tuning. However, the gap remains substantial even in the Full setups.

\begin{table}[!t]
\def\arraystretch{0.7}
\centering
{\scriptsize
\begin{tabularx}{\columnwidth}{l XXX}
\toprule
{\textbf{Variant}} & {\bf 10} & {\bf 30} & {\bf Full} \\
\cmidrule(lr){2-4}
\textsc{rob+s1+s2-cos} & {\bf 82.37} & {\bf 94.39} & {\bf 98.12} \\
\textsc{rob+s2-cos} & \underline{70.71} & \underline{92.14} & {97.42} \\
\cmidrule(lr){2-4}
\rowcolor{Gray}
\multicolumn{4}{c}{\bf MLP-Based} \\
\cmidrule(lr){2-4}
\textsc{rob+s1} & {48.26} & {85.49} & {97.16} \\
\textsc{USE} & {47.25} & {87.21} & {97.31} \\
\textsc{LaBSE} & {43.10} & {87.32} & \underline{97.42} \\
\bottomrule
\end{tabularx}
}%
\vspace{-1.5mm}
\caption{Results on English ATIS (Accuracy $\times 100$).}
\label{tab:atis}
\vspace{-2.5mm}
\end{table}

Figure~\ref{fig:rand-main} also reveals that the strength of \convfit Stage 1 is in adapting the knowledge acquired at LM pretraining: S1 fine-tuning of \textsc{rand} with smaller amounts of Reddit data cannot match \rob as the input LM, although the gap does become smaller with more in-task data for S2.

\vspace{1.3mm}
\noindent \textbf{Stage 2: Fine-Tuning Losses.} Table~\ref{tab:main} reveals that strong ID performance after S2 tuning is achieved with different loss functions from \S\ref{ss:stage2}, with different input LMs, even without any careful tuning of hyper-parameters for single settings. This verifies the versatility and robustness of \convfit. Both \cosine and \nce yield consistently strong results, and we expect that even higher absolute scores might be achieved by applying more sophisticated (contrastive learning) loss functions from prior work \cite{Hermans:2017arxiv,Liu:2020arxiv} in Stage 2.

\subsection{Further Discussion}
\label{ss:further}

\noindent \textbf{Stage 1: Amount of Reddit Examples.}
We now analyse what amount of Reddit data is required to turn input LMs into conversational encoders, by reducing S1 fine-tuning data through subsampling. The scores over different sizes are provided in Figure~\ref{fig:sizes-ocl}, and we note that they extend to other \convfit variants (see \S\ref{ss:variants}). As expected, having more Reddit data does yield better results on average, but even a small sample of Reddit data (e.g., $\approx$50K $(c,r)$ pairs) \textbf{1)} transforms the input LM into an effective sentence encoder (e.g., its MLP-based ID results are on par with those achieved with USE, LaBSE, and ConveRT), and \textbf{2)} improves over the \convfit variant that skips S2 completely. This implies that perhaps more careful domain-driven data sampling in the future might yield even more domain-adapted conversational encoders after S1.

\vspace{1.3mm}
\noindent \textbf{Amount of Negative Examples} in Stage 2 has only a moderate to negligible impact on the final performance, as shown in Figure~\ref{fig:neg-ocl}. Small gains when moving from $n=1$ to $n=3$ are observed only for the 10-shot setup: there, having more negatives may implicitly play the role of data augmentation for fine-tuning. However, with more in-task examples, the dependence on $n$ becomes inconsequential, and the performance saturates quickly (e.g., see the curves in the 30-shot setups).


\vspace{1.3mm}
\noindent \textbf{Stage 2: Few-Shot versus Full.} Framing the ID task a sentence similarity seems especially beneficial for few-shot scenarios, as the model can leverage prototype-based (or instance-based) similarities \cite{Snell:2017nips} in the specialised encoder space. However, the strong performance with fully \convfit-ed models persists also in Full setups. This finding is further corroborated with the results on another standard ID dataset, English ATIS \cite{Hemphill:1990,Xu:2020emnlp}, see Table~\ref{tab:atis}. There, we observe even more prominent differences in favour of similarity-based ID enabled by \convfit, again especially in the two low-data setups. The proposed prototype-based learning and inference holds promise to boost few-shot performance even more in future work, through additional metric learning \cite{Zhang:2020emnlp} or data augmentation techniques \cite{Lee:2021arxiv}.

\begin{figure}[!t]
    \centering
    \includegraphics[width=0.87\columnwidth]{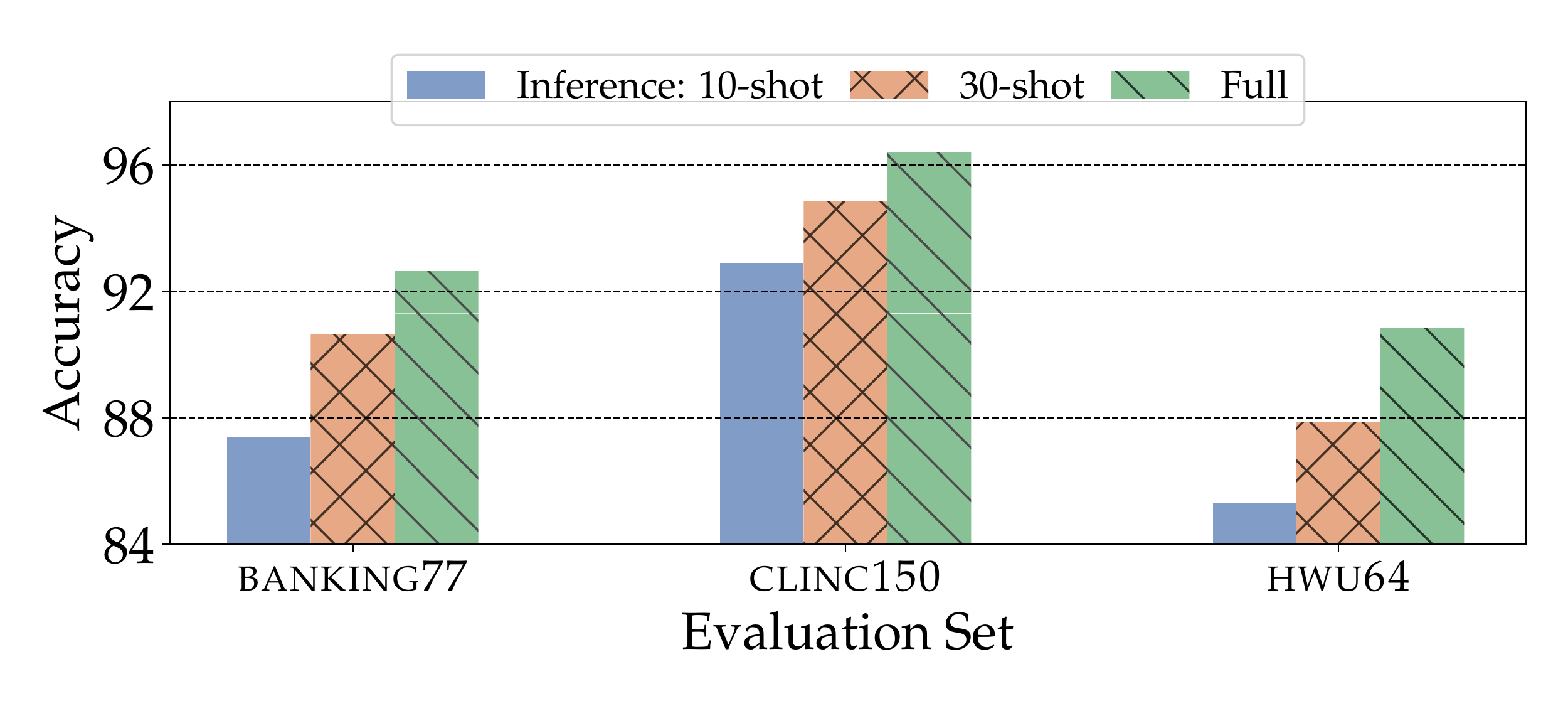}
    \vspace{-4mm}
    \caption{Impact of the number of data instances at inference. The \textsc{rob+s1+s2-ocl} variant is tuned in 10-shot setups in S2, and additional data (30-shot or Full) is used only at inference without any S2 retuning.}
    \vspace{-3.5mm}
    \label{fig:infer-main}
\end{figure}

One limitation of \convfit, especially prominent in Full scenarios, is its quadratic time complexity. Future work will look into effective sampling strategies and adaptations towards more sample-efficient and quicker fine-tuning \cite{Tran:2019sigir,Tian:2020iclr,ONeill:2021arxiv}. 

\vspace{1.3mm}
\noindent \textbf{Data Augmentation for Inference.}
Adding more data instances for similarity-based inference, serving as exemplars/prototypes, is likely to boost the final intent detection performance \textit{without the need to retrain the model}. The intuition is that additional instances can provide finer-grained prototypes for inference, semantically more similar to the input query sentences than the original training data. To test this hypothesis, we conduct a simple probing experiment, where we train the \textsc{rob+s1+s2-ocl} ($n=3$) variant in the 10-shot setup, but then run inference (i) with the same 10 shots; (ii) in the 30-shot setup (i.e., effectively performing the inference-time data augmentation, relying on 20 more data instances per intent class at inference); (iii) in the Full setup. 

The scores are summarised in Figure~\ref{fig:infer-main}. They clearly indicate that performance does rise with more data instances at inference, even without any model retraining/re-tuning, confirming that increased semantic variability helps at inference. This finding is salient for all three evaluation sets.\footnote{The same trends persist with other \convfit variants.} As expected, the absolute performance of 30-shot or Full inference when the model is trained in 10-shot setups is lower than in the setup where the more abundant data is additionally used for \convfit Stage 2 task-tuning. 

Based on these findings, we restate that a promising path for future research concerns investigating and `task-adapting' automatic paraphrase generation models \cite{Krishna:2020emnlp,Dopierre:2021acl,Schick:2021arxiv} such as the one that rely on prompting large models (e.g., GPT-3, T5) \cite{Gao:2021acl}. Such paraphrases might provide a richer and semantically more varied set of data instances for \convfit task-tailored fine-tuning and similarity-based inference.



%% file: 05-conclusion.tex

We proposed \convfit, a two-stage \textit{conversational fine-tuning} procedure that transforms pretrained LMs (e.g., BERT, RoBERTa) into universal (after Stage 1) and task-specialised conversational sentence encoders (after Stage 2) through dual-encoder architectures. The semantic knowledge already stored in the pretrained LMs gets 'rewired' for a particular domain and task. We demonstrated that such task-specialised sentence encoders enable casting intent detection (ID) as simple sentence similarity;  \convfit-ed encoders yield strong ID results across diverse ID datasets and setups.

The \convfit framework is very versatile and opens up many future research paths and further extensions and experimentation beyond the scope of this paper. For instance, it is possible to replace the current contrastive loss functions with other recent effective contrastive losses \cite[\textit{inter alia}]{Oord:2018arxiv,Gunel:2021iclr}, or mine hard (instead of using random) negative examples \cite{Lauscher:2020coling,Kalantidis:2020neurips,Robinson:2021iclr}. We will also extend \convfit to other pretrained models, experiment with automatic paraphrasers for data augmentation, and port the framework to other conversational tasks (e.g., slot labelling for dialogue), as well as to other, non-dialogue text classification tasks.

%% file: xx-appendix.tex
\section{Additional Experiments and Results}

Additional experiments and analyses that further support the main claims of the paper have been relegated to the appendix for clarity and compactness of our presentation in the main paper. They largely follow the trends observed in the results which are provided in the main paper. In sum, we provide the following additional results and information, which offer further empirical support of our main claims in this paper:

\vspace{1.5mm}
\noindent \textbf{Table~\ref{tab:main-app}} provides the results with all input LMs in our comparison in all the \convfit variants discussed in \S\ref{ss:variants} across different data setups on all three intent detection datasets. It can be seen as a full (i.e., expanded) version of Table~\ref{tab:main} provided in the main paper.

\begin{figure*}[!t]
    \centering
    \begin{subfigure}[!ht]{0.326\linewidth}
        \centering
        \includegraphics[width=0.999\linewidth]{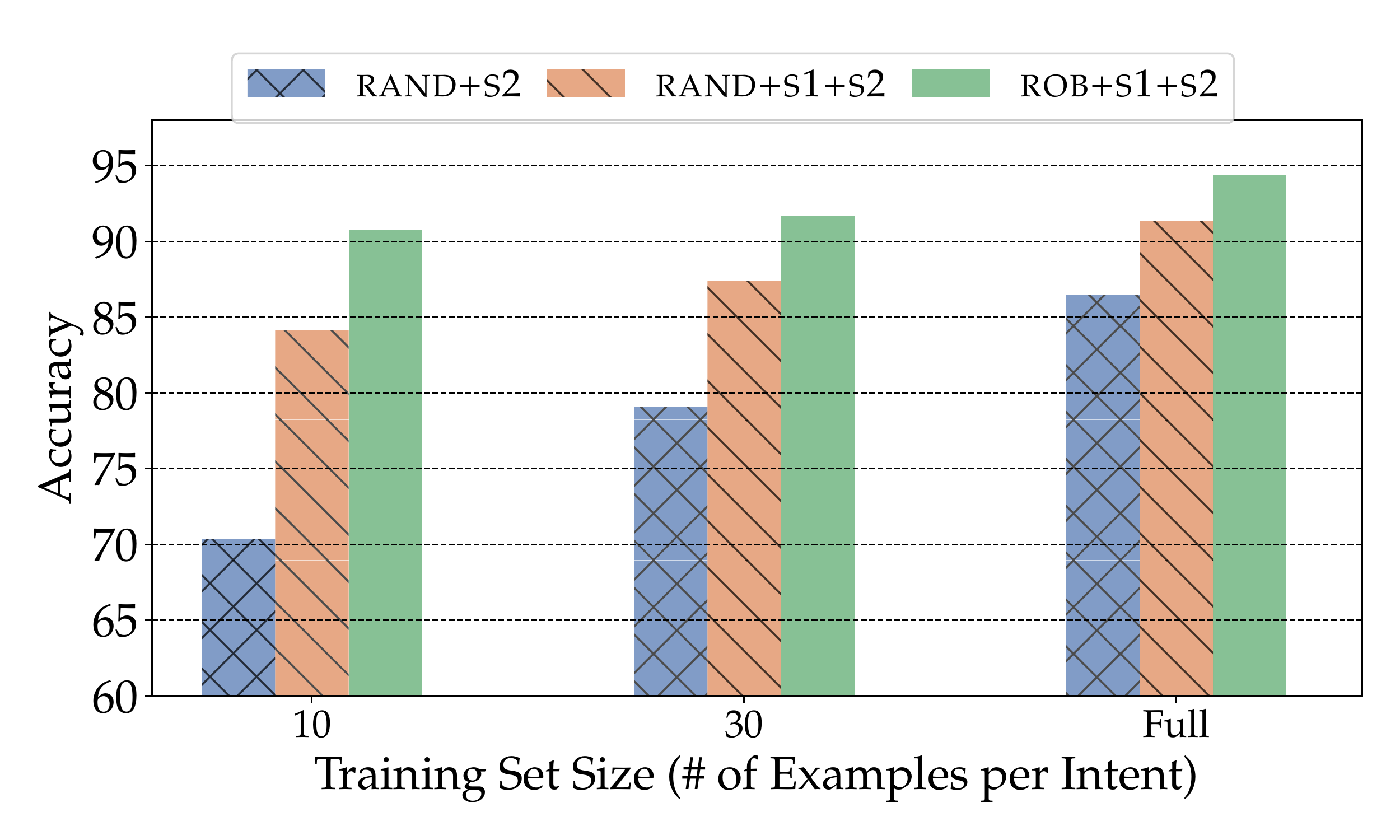}
        \caption{\banking}
        \label{fig:rand-cos-banking}
    \end{subfigure}
        \begin{subfigure}[!ht]{0.326\linewidth}
        \centering
        \includegraphics[width=0.999\linewidth]{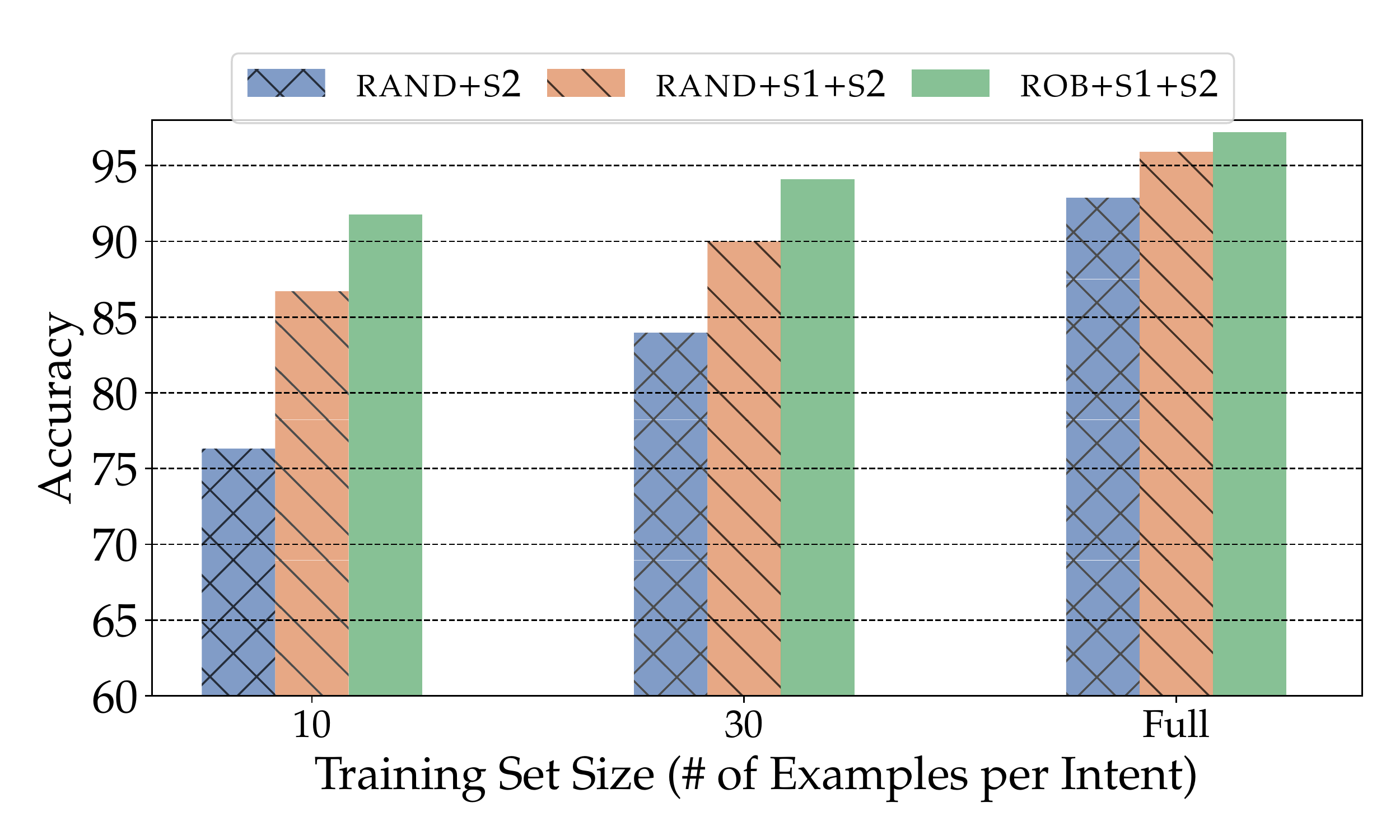}
        \caption{\clinc}
        \label{fig:rand-cos-clinc}
    \end{subfigure}
    \begin{subfigure}[!ht]{0.326\linewidth}
        \centering
        \includegraphics[width=0.999\linewidth]{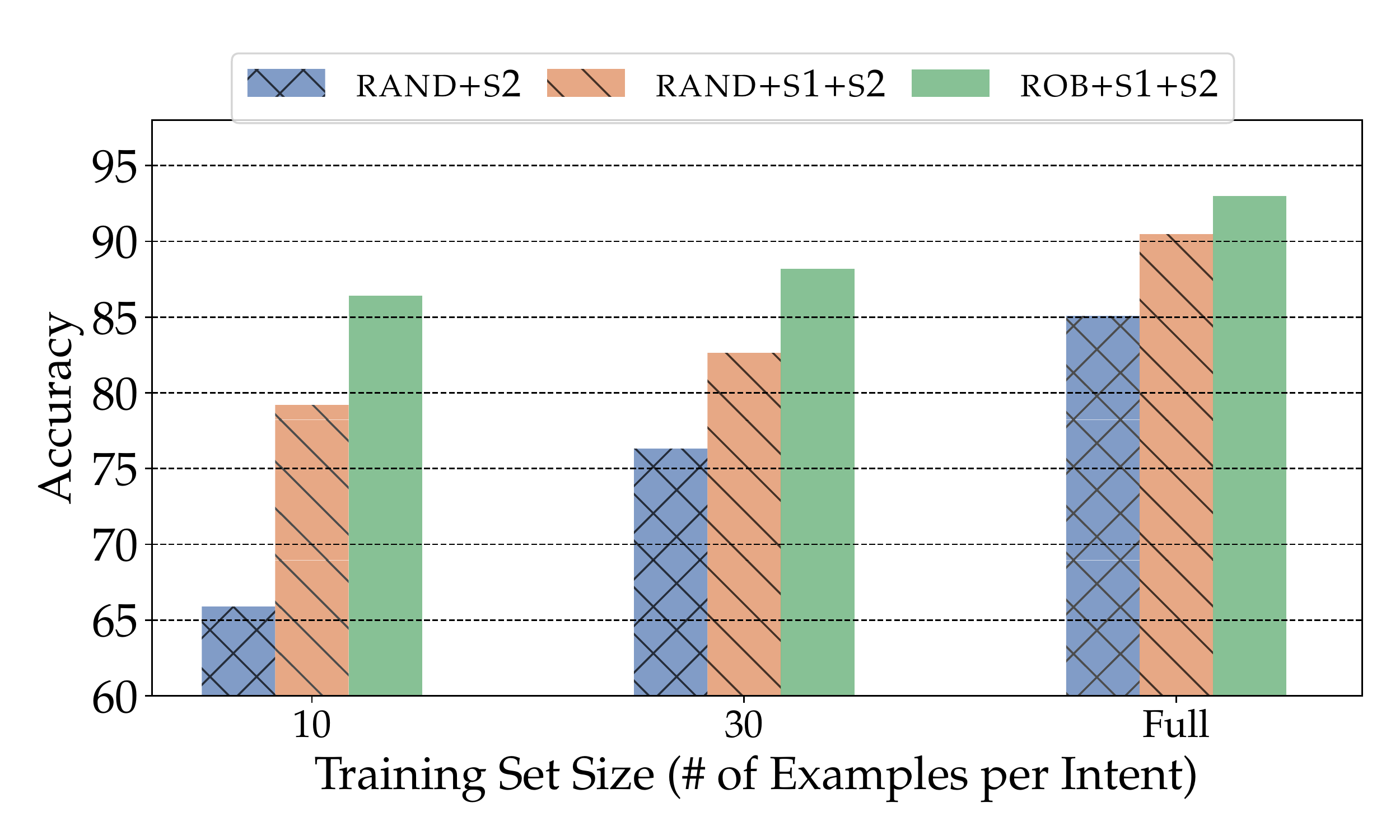}
        \caption{\hwu}
        \label{fig:rand-cos-hwu}
    \end{subfigure}
    \vspace{-1.5mm}
    \caption{A comparison of a randomly initialized BERT or RoBERTa architecture (\textsc{rand}) with LM-pretrained RoBERTa after Stage 2 \convfit-ing; evaluation on all three intent detection datasets; the \cosine loss used in S2. Figure~\ref{fig:rand-ocl-app} shows the similar plots with the \nce loss used in S2.}
    \vspace{-1.5mm}
\label{fig:rand-cos-app}
\end{figure*}

\begin{figure*}[!t]
    \centering
    \begin{subfigure}[!ht]{0.326\linewidth}
        \centering
        \includegraphics[width=0.999\linewidth]{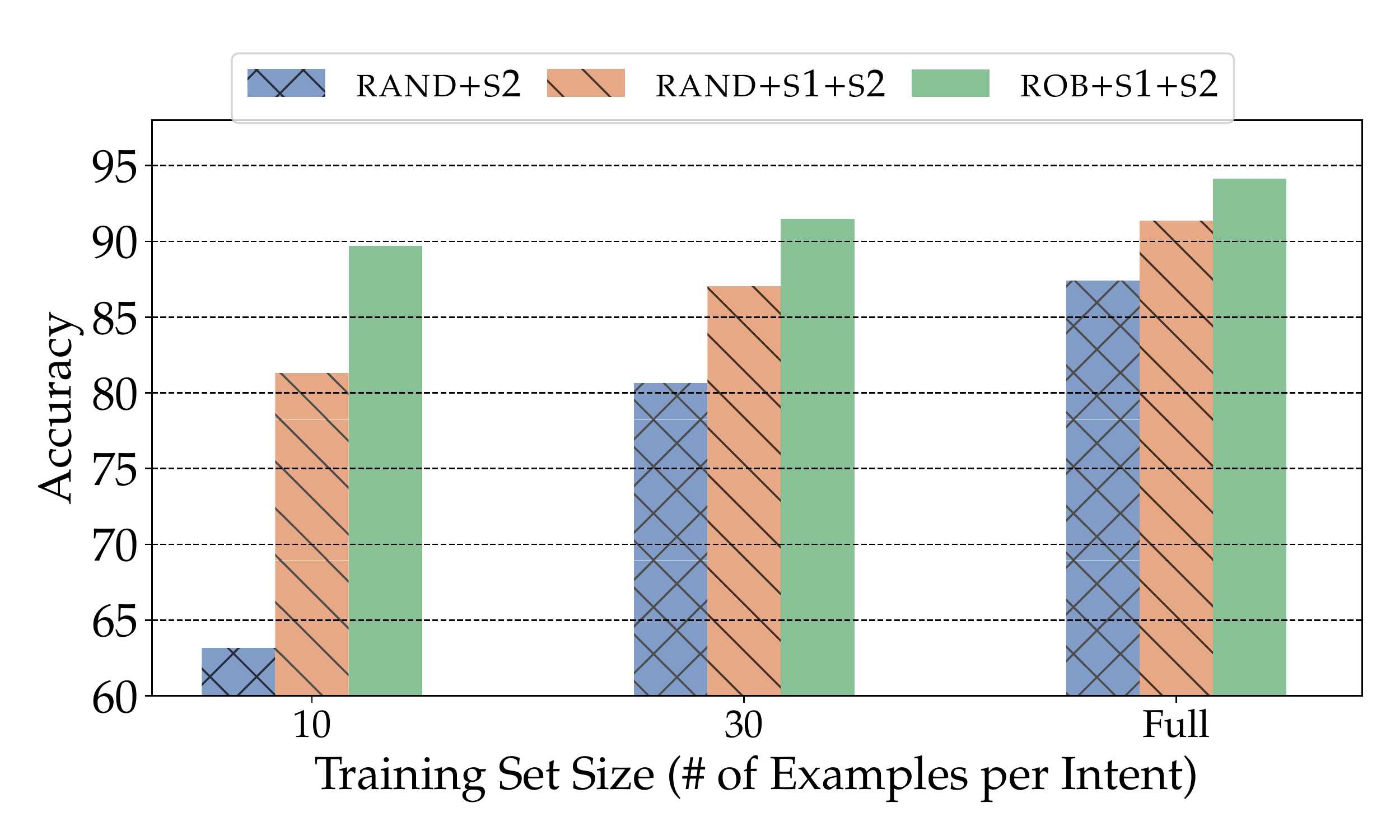}
        \caption{\banking}
        \label{fig:rand-ocl-banking}
    \end{subfigure}
        \begin{subfigure}[!ht]{0.326\linewidth}
        \centering
        \includegraphics[width=0.999\linewidth]{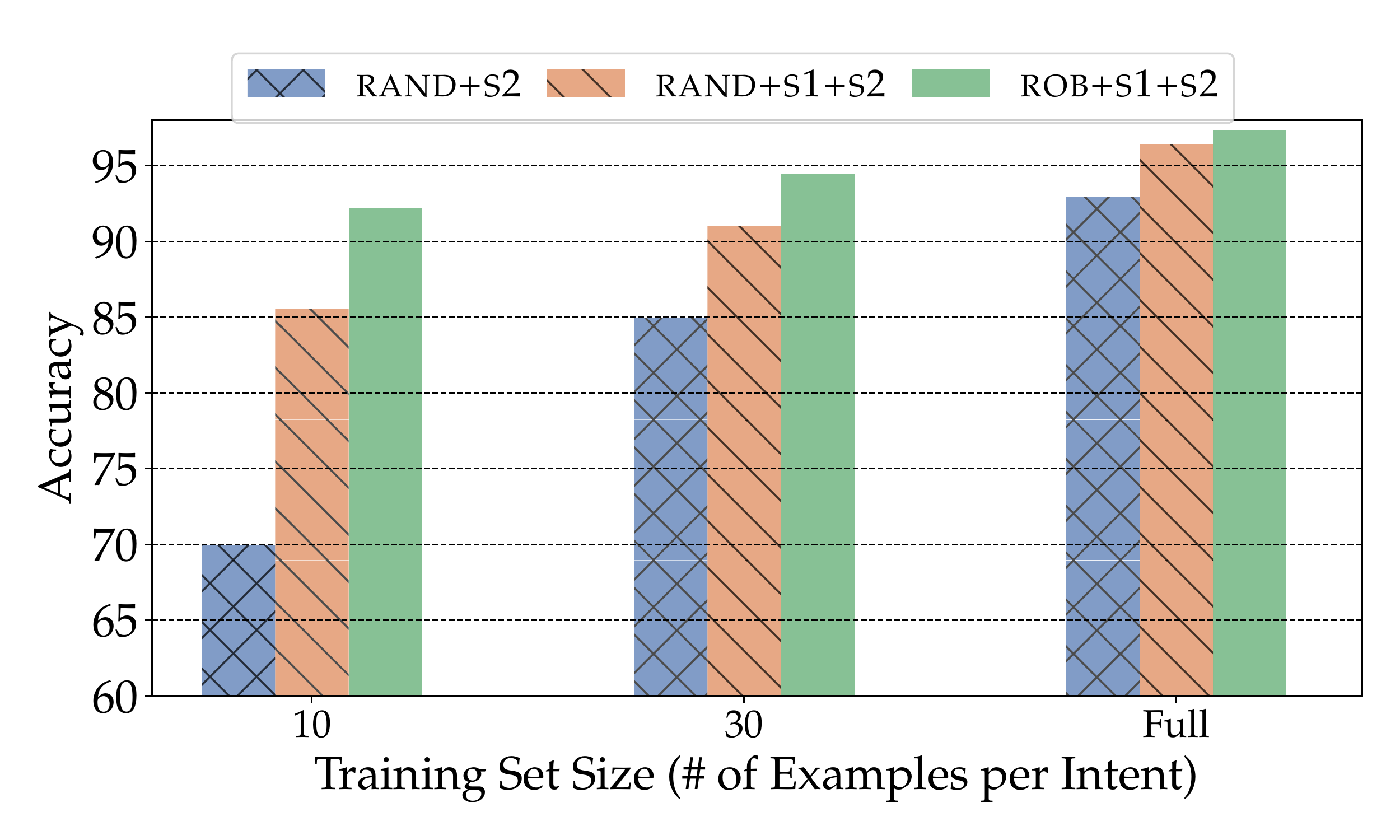}
        \caption{\clinc}
        \label{fig:rand-ocl-clinc}
    \end{subfigure}
    \begin{subfigure}[!ht]{0.326\linewidth}
        \centering
        \includegraphics[width=0.999\linewidth]{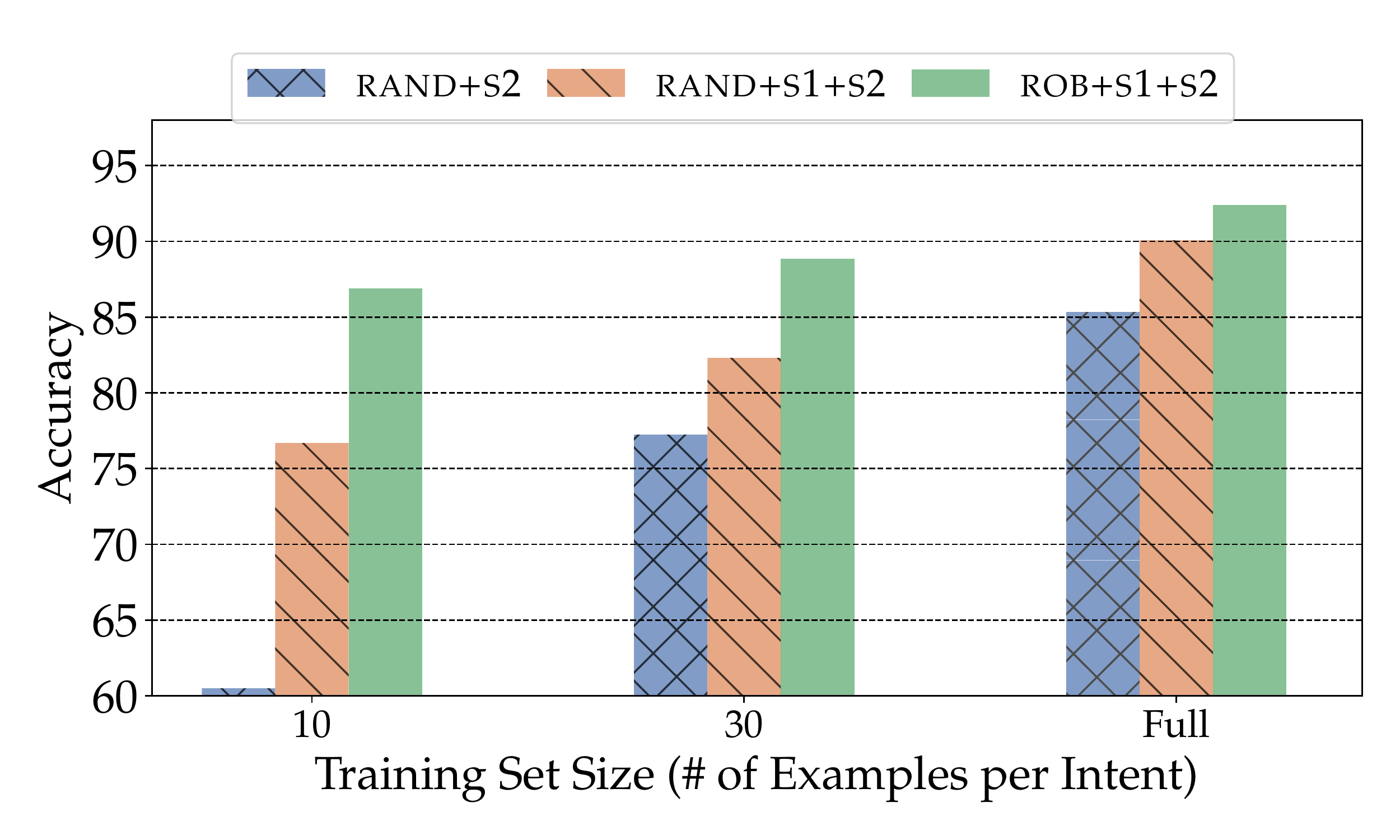}
        \caption{\hwu}
        \label{fig:rand-ocl-hwu}
    \end{subfigure}
    \vspace{-1.5mm}
    \caption{A comparison of a randomly initialized BERT or RoBERTa architecture (\textsc{rand}) with LM-pretrained RoBERTa after Stage 2 \convfit-ing; evaluation on all three intent detection datasets; the \nce loss used in S2.}
    \vspace{-1.5mm}
\label{fig:rand-ocl-app}
\end{figure*}

\vspace{1.5mm}
\noindent \textbf{Figure~\ref{fig:rand-cos-app}} (\cosine loss in Stage 2) and \textbf{Figure~\ref{fig:rand-ocl-app}} (\nce loss in Stage 2) demonstrate the impact of using LM-pretrained Transformers versus randomly initialised Transformers in the \convfit framework (both in the full S1+S2 setup, as well as in the setup where only task-tuning (S2) is employed). The patterns in the results, presented over all three evaluation sets, largely follow the patterns observed in Figure~\ref{fig:rand-main}, which is provided in the main paper.

\begin{figure*}[t!]
    \centering
    \begin{subfigure}[!ht]{0.464\linewidth}
        \centering
        \includegraphics[width=0.98\linewidth]{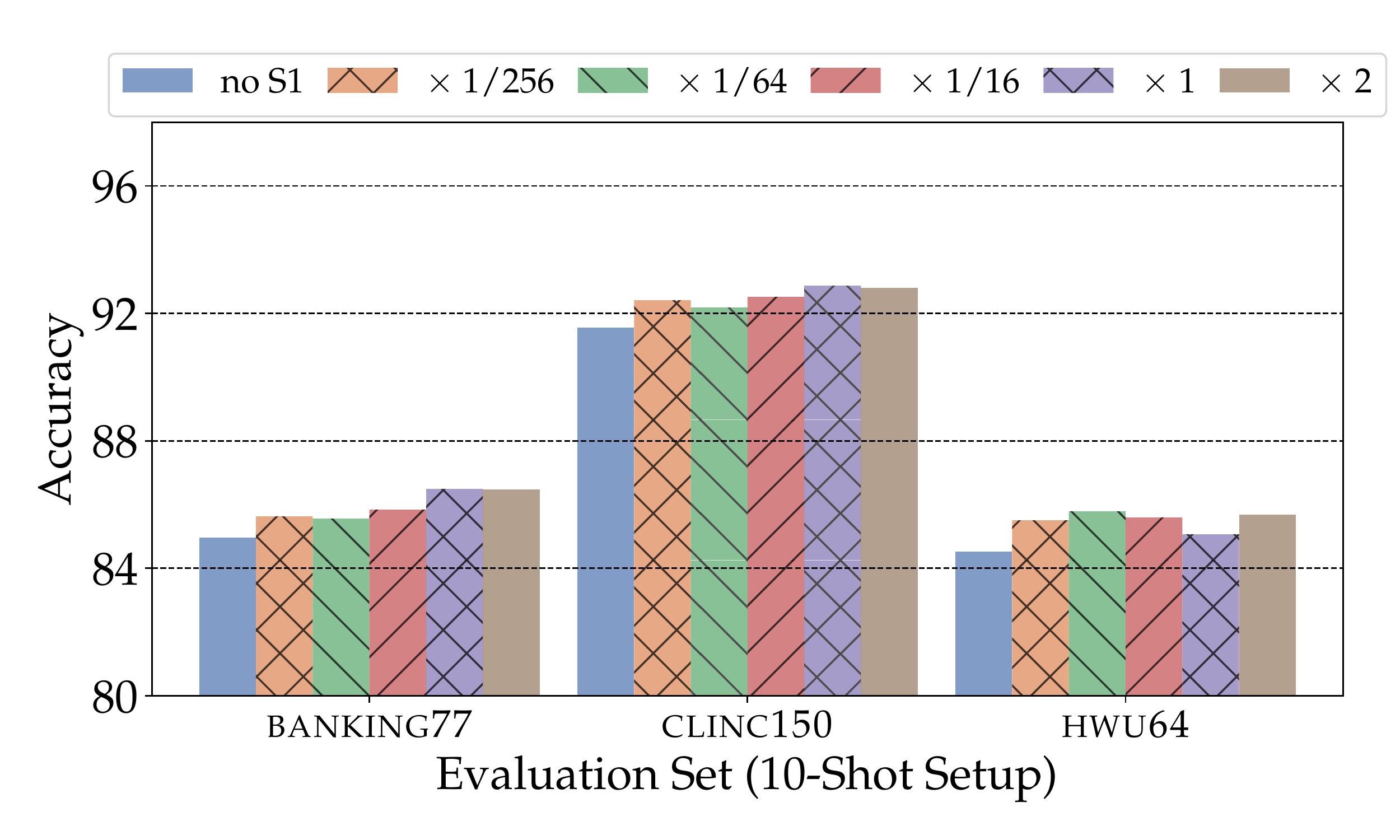}
        \caption{10-shot (\textsc{rob+s1+s2-cos})}
        \label{fig:sizes-cos-10}
    \end{subfigure}
    \begin{subfigure}[!ht]{0.464\textwidth}
        \centering
        \includegraphics[width=0.98\linewidth]{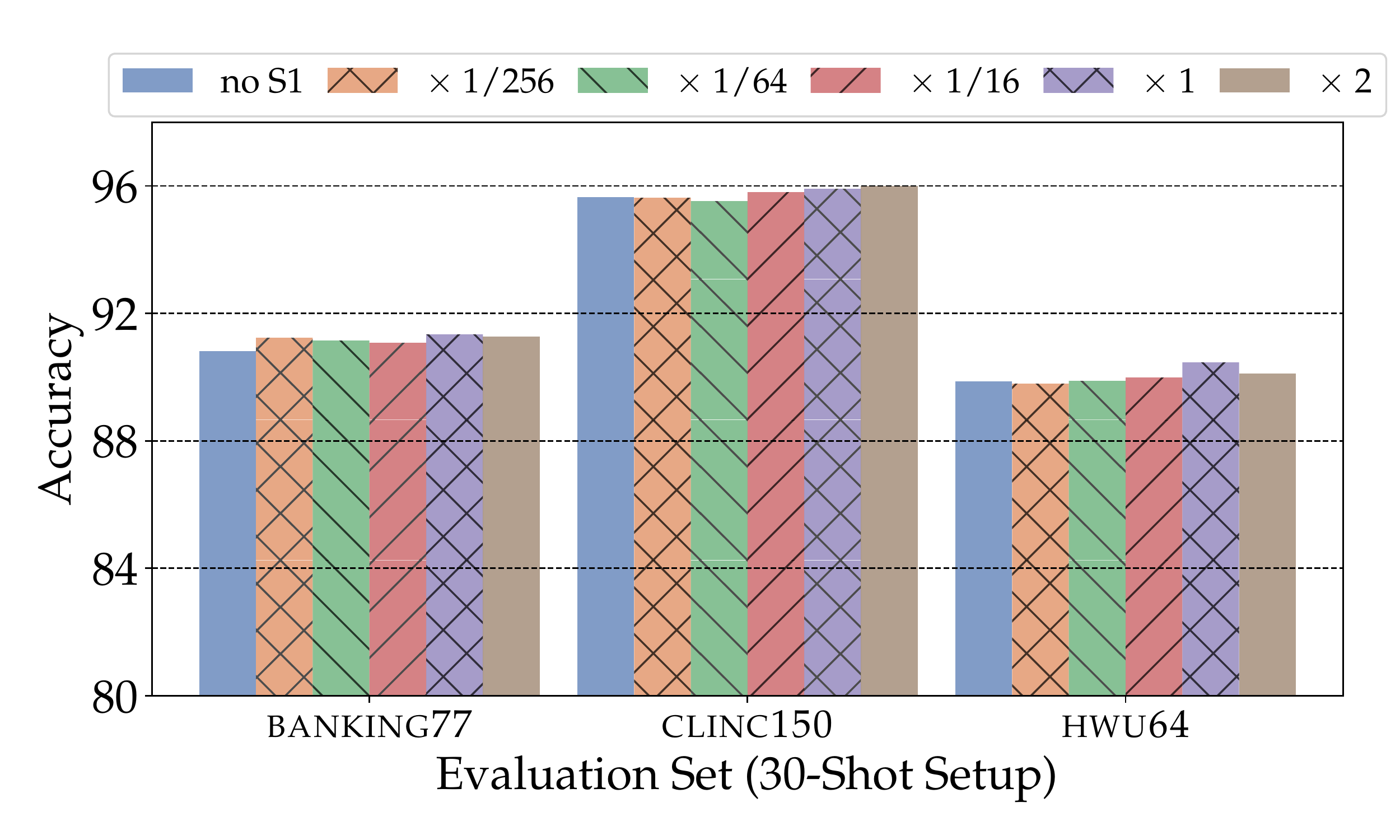}
        \caption{30-shot (\textsc{rob+s1+s2-cos})}
        \label{fig:sizes-cos-30}
    \end{subfigure}
    \vspace{-1.5mm}
    \caption{Varying the amount of Reddit data for Stage 1 \convfit; $\times 1$ refers to the Reddit size used in all our other Stage 1 fine-tuning experiments ($\approx$2\% of the full Reddit corpus from \newcite{Henderson:2019arxiv}), while other Reddit data sizes are relative to this corpus size (e.g., $\times 1/32$ means that we use $2\%/32\approx0.0625\%$ of the full Reddit corpus). Stage 2 loss is \cosine ($n=3$).}
    \vspace{-1.5mm}
\label{fig:sizes-cos}
\end{figure*}

\vspace{1.5mm}
\noindent \textbf{Figure~\ref{fig:sizes-cos}} plots how the amount of Reddit data in Stage 1 impacts the final intent detection performance when the \cosine loss is used for task-tuning in Stage 2. The observed trends in results are very similar to the ones obtained with the \nce loss, presented in the main paper (see Figure~\ref{fig:sizes-ocl}). 

\begin{figure*}[!t]
    \centering
    \begin{subfigure}[!ht]{0.326\linewidth}
        \centering
        \includegraphics[width=0.999\linewidth]{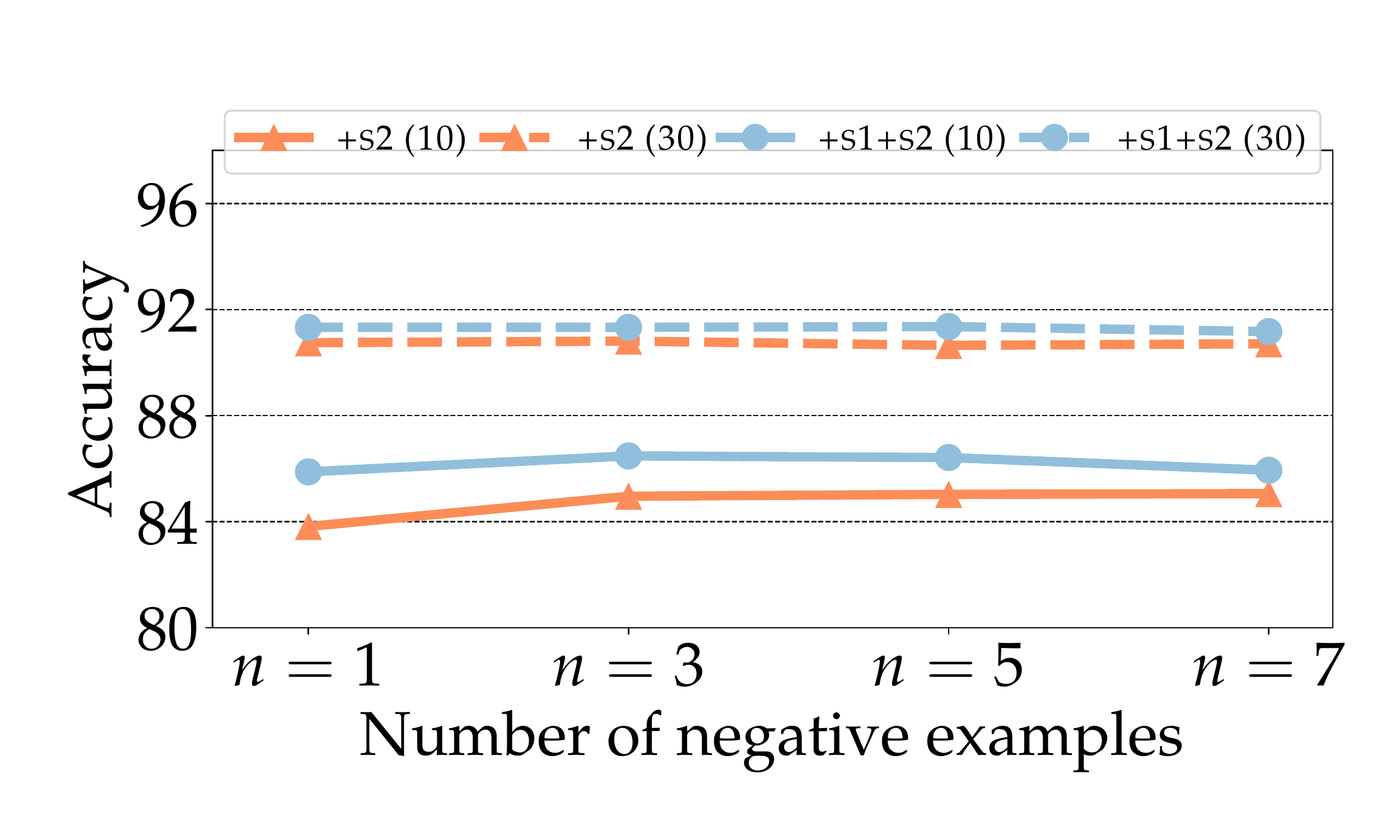}
        \caption{\banking}
        \label{fig:neg-cos-banking}
    \end{subfigure}
        \begin{subfigure}[!ht]{0.326\linewidth}
        \centering
        \includegraphics[width=0.999\linewidth]{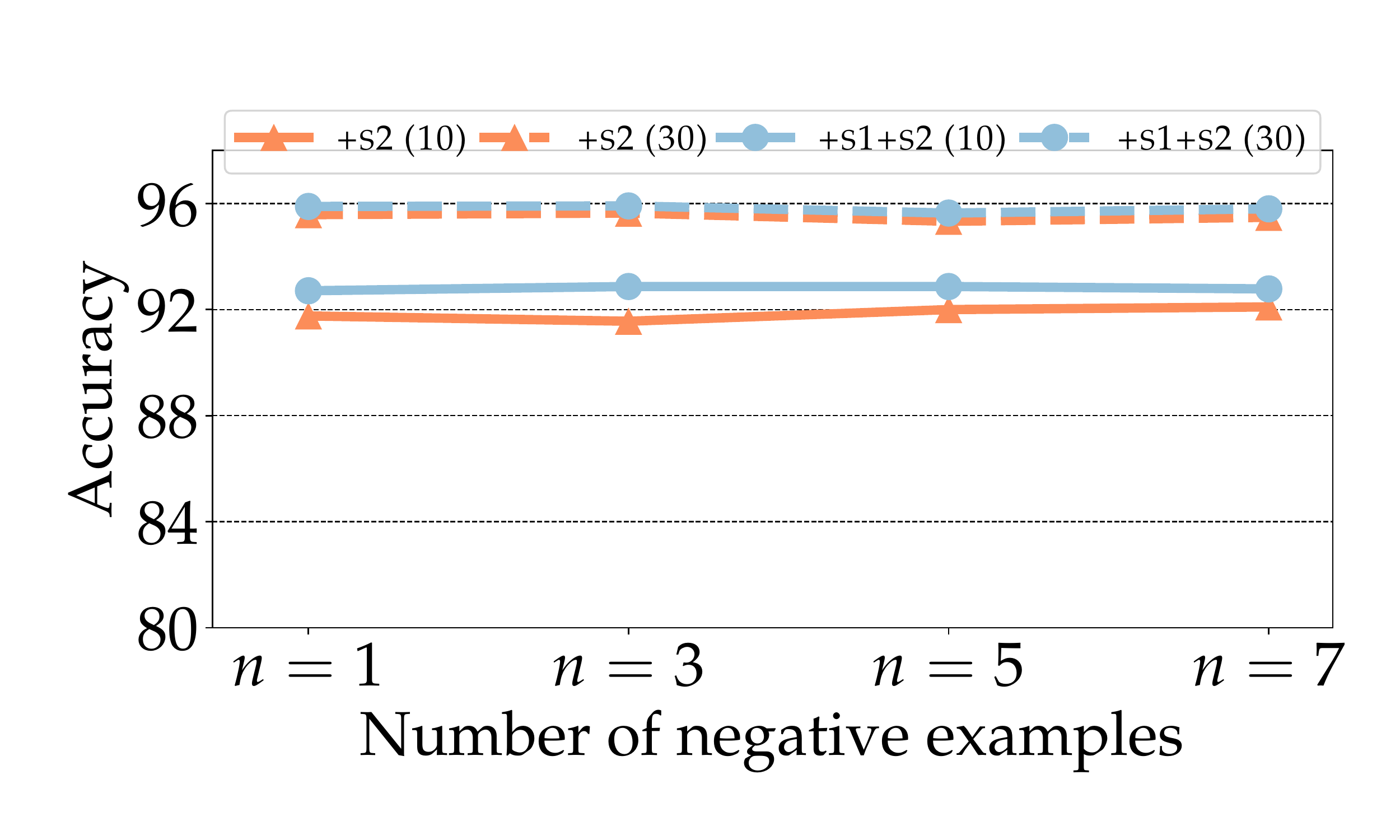}
        \caption{\clinc}
        \label{fig:neg-cos-clinc}
    \end{subfigure}
    \begin{subfigure}[!ht]{0.326\linewidth}
        \centering
        \includegraphics[width=0.999\linewidth]{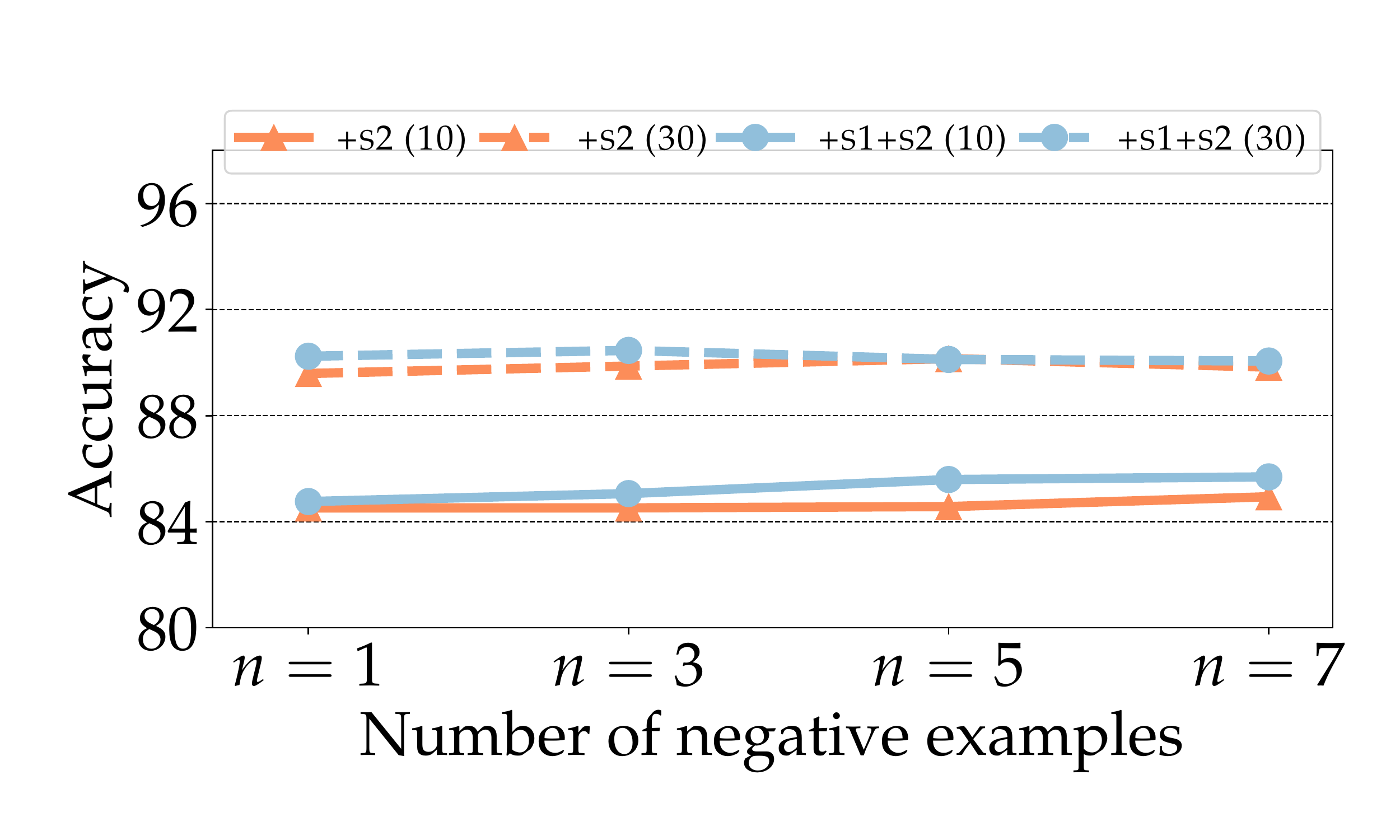}
        \caption{\hwu}
        \label{fig:relp-neg-cos-hwu}
    \end{subfigure}
    \vspace{-1.5mm}
    \caption{Impact of the number of negative examples $n$ on intent detection performance in 10-shot and 30-shot setups. The \convfit model variants are \textsc{rob+s2+cos} and \textsc{rob+s1+s2+cos}, that is, RoBERTa is the input LM in all experiments, and the results show model variants with the \cosine loss in Stage 2, without and with S1 fine-tuning (labelled +S2 and +S1+S2 in the figures, respectively).}
    \vspace{-1.5mm}
\label{fig:neg-cos}
\end{figure*}

\vspace{1.5mm}
\noindent \textbf{Figure~\ref{fig:neg-cos}} presents the impact of the number of negative examples $n$ during Stage 2 fine-tuning with the \cosine loss; the observed trends are very similar to the ones with the \nce loss, presented in the main paper (see Figure~\ref{fig:neg-ocl}).

\begin{figure}[b]
    \centering
    \includegraphics[width=0.8\columnwidth]{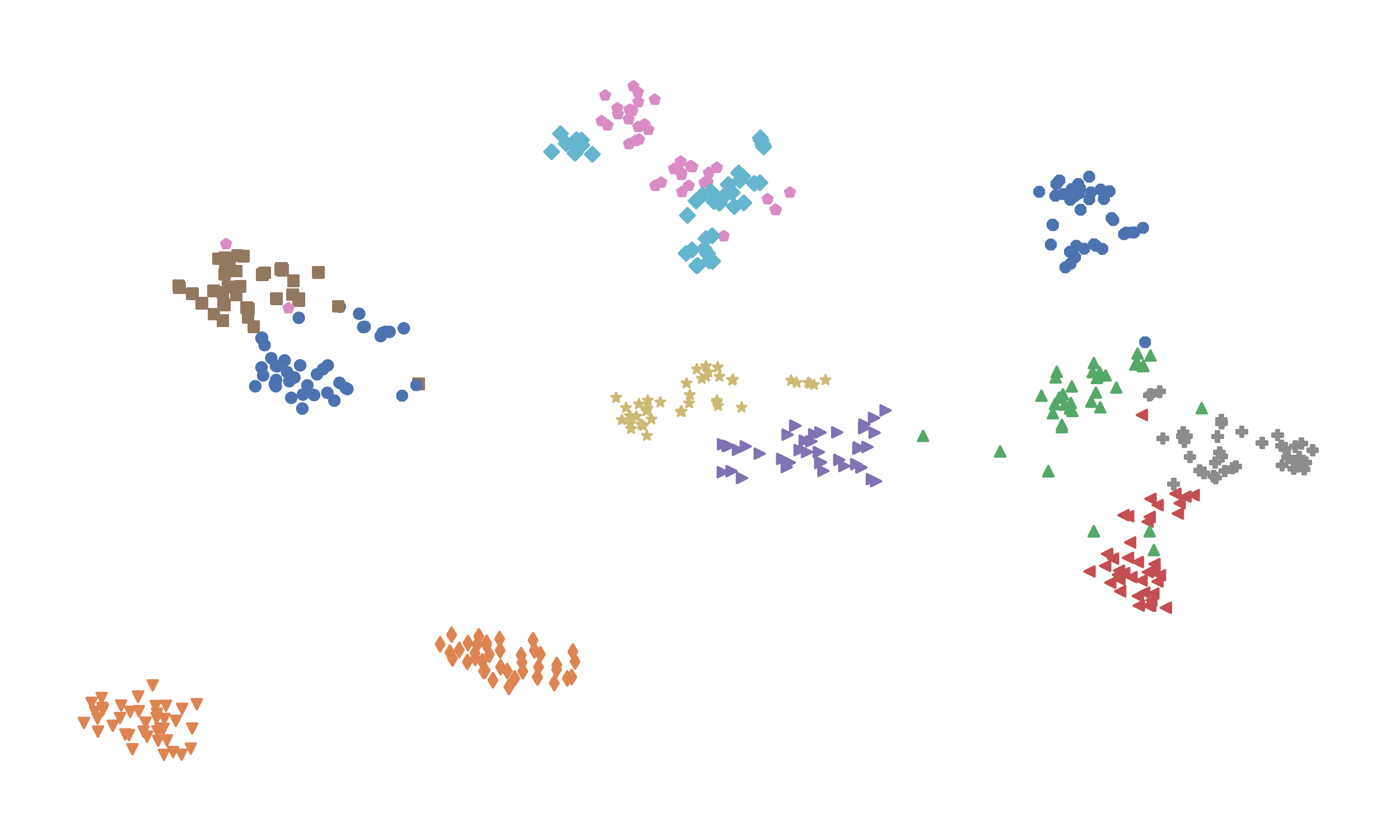}
    \vspace{-2mm}
    \caption{t-SNE plots of encoded utterances from the test set of \textsc{banking77} (a subset of 12 intents, see the legend in Figure~\ref{fig:tsne-rob}) after Stage 1 fine-tuning of RoBERTa using only $\approx$50k \textit{(context, response)} pairs from Reddit; cf., Figure~\ref{fig:tsne-orig}.}
    \label{fig:tsne-r50k}
\end{figure}

\begin{figure*}[p]
    \centering
    \begin{subfigure}[!ht]{0.98\textwidth}
        \centering
        \includegraphics[width=1.0\linewidth,trim=0cm 0cm 0cm 0.1cm,clip]{./Figures/tsne-legend}
        \label{fig:legend2}
        \vspace{-3.5mm}
    \end{subfigure}
    \begin{subfigure}[!ht]{0.323\linewidth}
        \centering
        \includegraphics[width=0.98\linewidth]{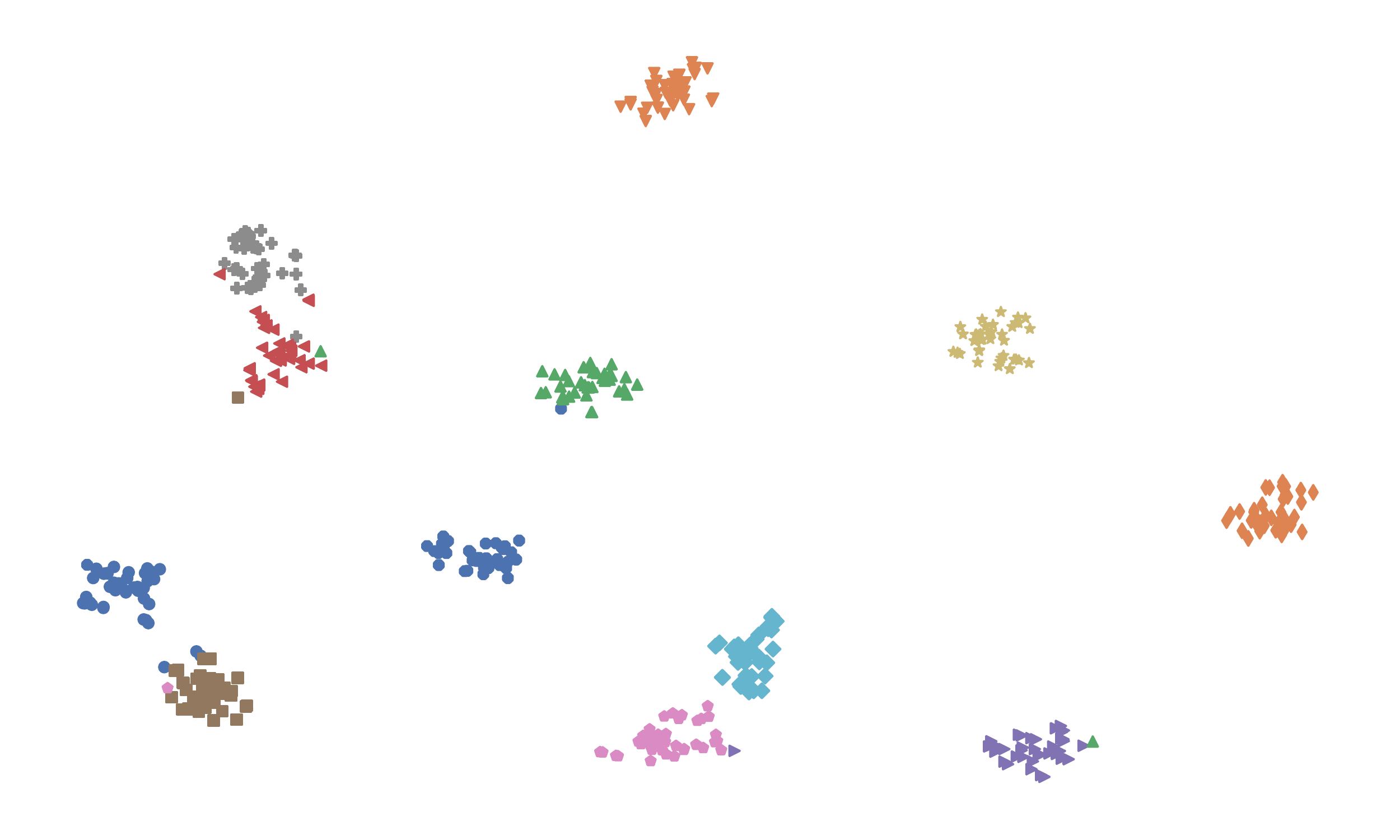}
        \caption{RoBERTa (10-shot S2)}
        \label{fig:rob-ocl-10}
    \end{subfigure}
    \begin{subfigure}[!ht]{0.323\textwidth}
        \centering
        \includegraphics[width=0.98\linewidth]{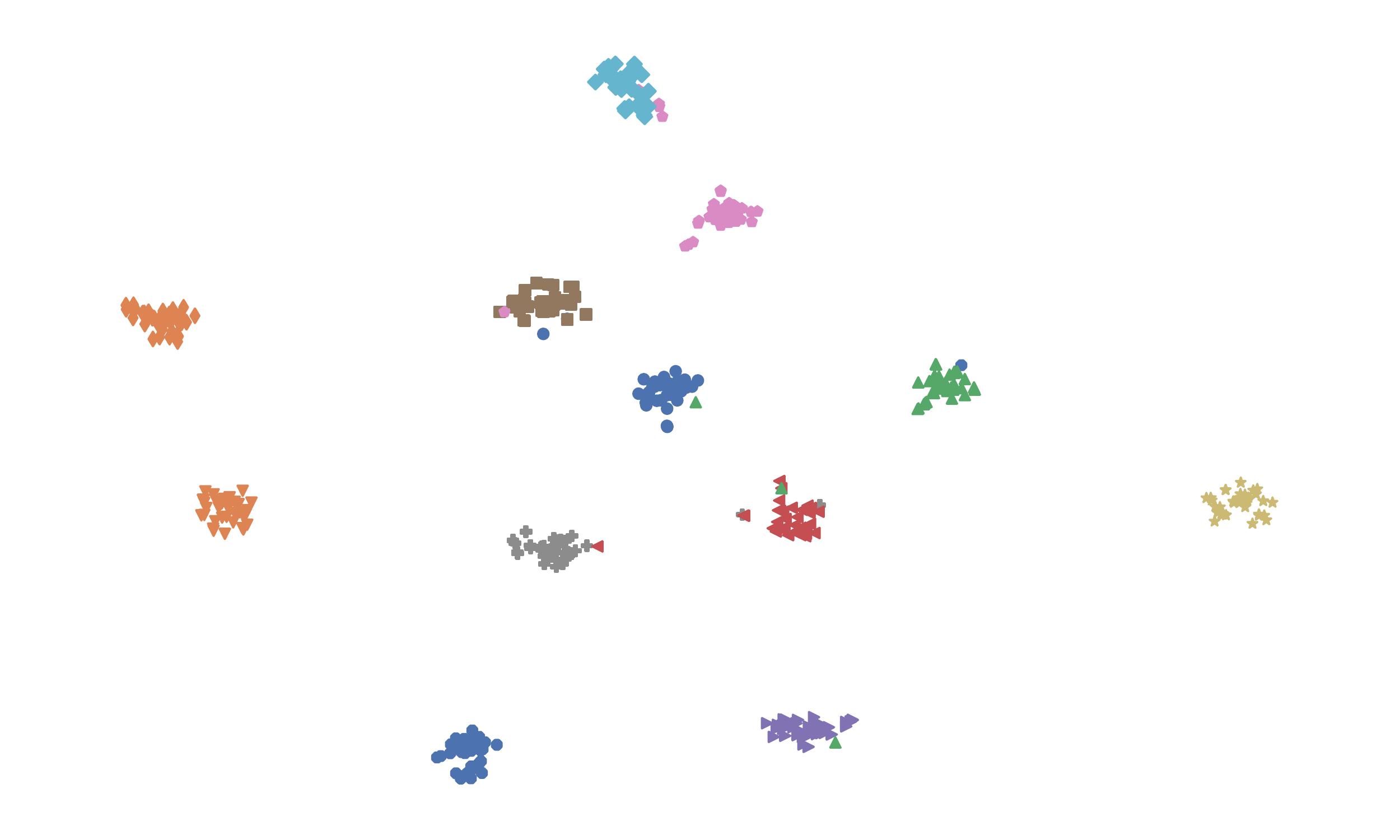}
        \caption{RoBERTa (30-shot S2)}
        \label{fig:rob-ocl-30}
    \end{subfigure}
    \begin{subfigure}[!ht]{0.323\linewidth}
        \centering
        \includegraphics[width=0.98\linewidth]{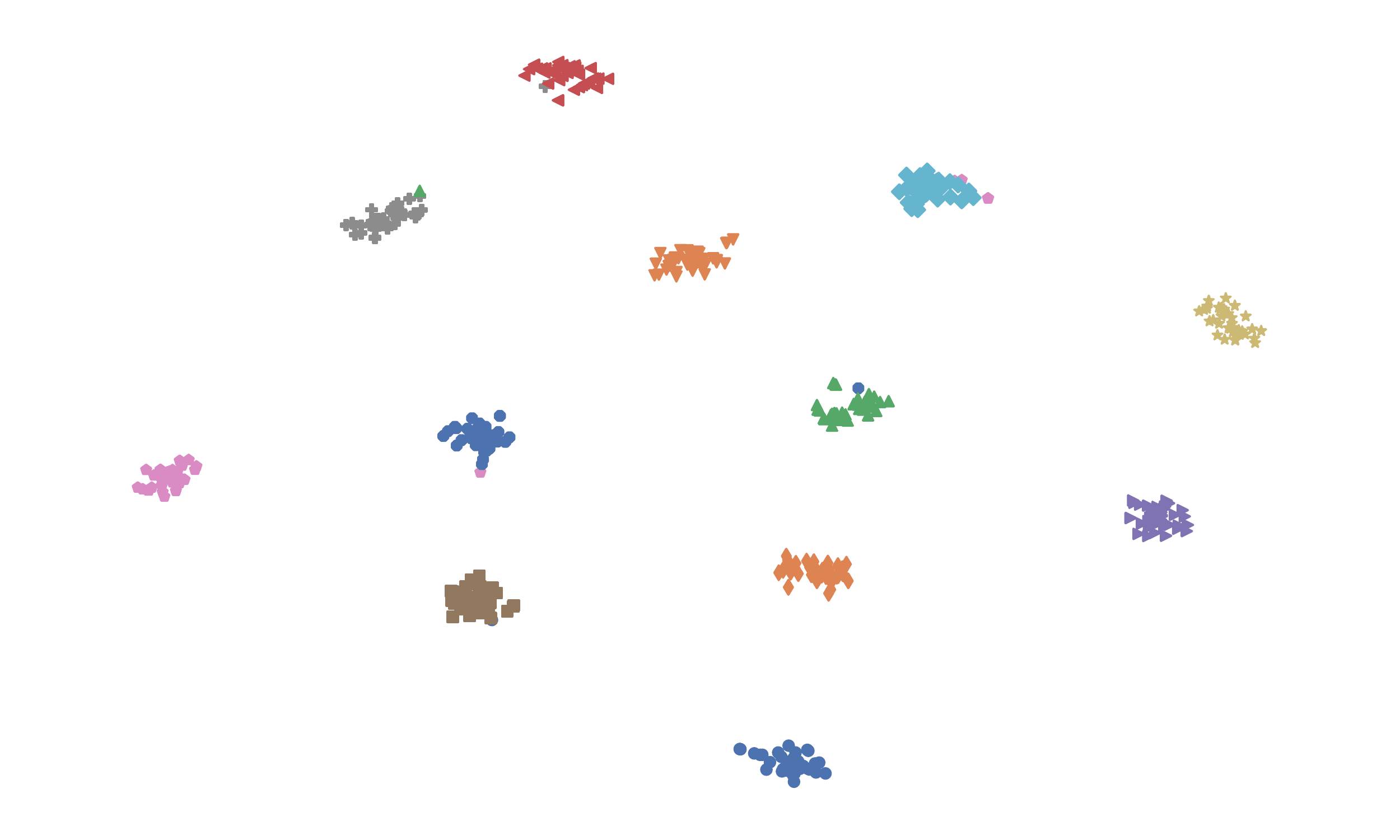}
        \caption{RoBERTa (Full S2)}
        \label{fig:rob-ocl-all}
    \end{subfigure}
    \vspace{-1.5mm}
    \caption{t-SNE plots \cite{tsne:2012} of encoded utterances from the test set of \textsc{banking77} (i.e., all examples are effectively unseen by the encoder models at training) associated with a selection of 12 intents. The encoded utterances are created via mean-pooling based on fine-tuned RoBERTa encoders which underwent Stage 1 plus Stage 2 in the \textbf{(a)} 10-shot Stage 2 setup (i.e., 10 examples per intent); \textbf{(b)} 30-shot setup; \textbf{(c)} Full setup (see also \S\ref{s:experimental}). Stage 2: fine-tuning with the \nce objective ($n=3$ negatives). The results suggest that even in 10-shot setups it is possible to learn coherent clusters and clear cluster separations; however, the clusters become less and less compact, and less separated in the semantic space as we fine-tune with fewer in-task instances (e.g., compare the clusters in the 10-shot versus Full setup), and the fine-tuned encoder model is more prone to incorrect cluster assignments. This (initially) visual observation is also supported by the Silhouette coefficient scores (higher is better): (a) $\sigma=0.378$, (b) $\sigma=0.548$, (c) $\sigma=0.698$.}
    \vspace{-1.5mm}
\label{fig:tsne-1030-rob}
\end{figure*}

\begin{figure*}[p]
    \centering
    \begin{subfigure}[!ht]{0.98\textwidth}
        \centering
        \includegraphics[width=1.0\linewidth,trim=0cm 0cm 0cm 0.1cm,clip]{./Figures/tsne-legend}
        \label{fig:legendxxx}
        \vspace{-3.5mm}
    \end{subfigure}
    \begin{subfigure}[!ht]{0.323\linewidth}
        \centering
        \includegraphics[width=0.98\linewidth]{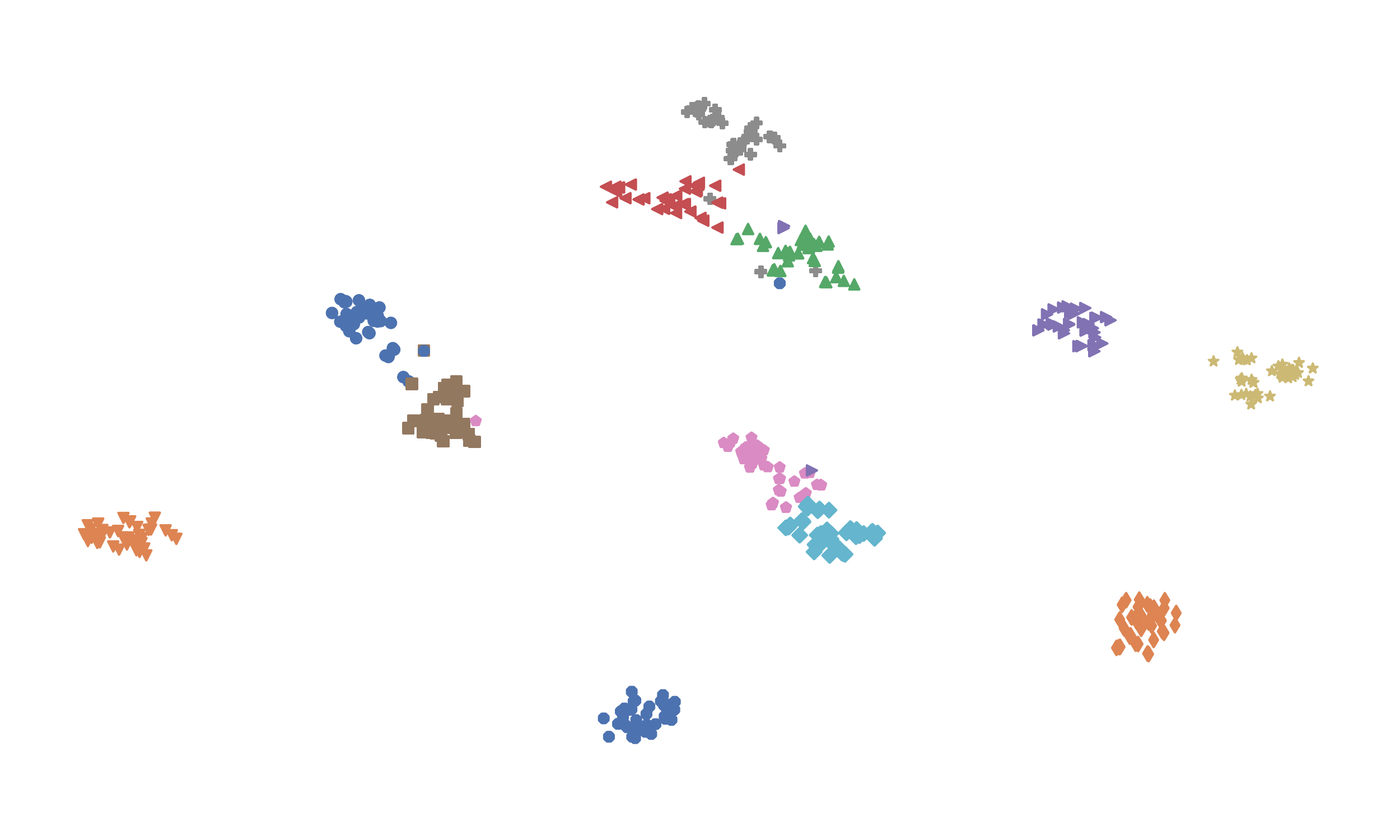}
        \caption{RoBERTa (10-shot S2)}
        \label{fig:rob-orig-10}
    \end{subfigure}
    \begin{subfigure}[!ht]{0.323\textwidth}
        \centering
        \includegraphics[width=0.98\linewidth]{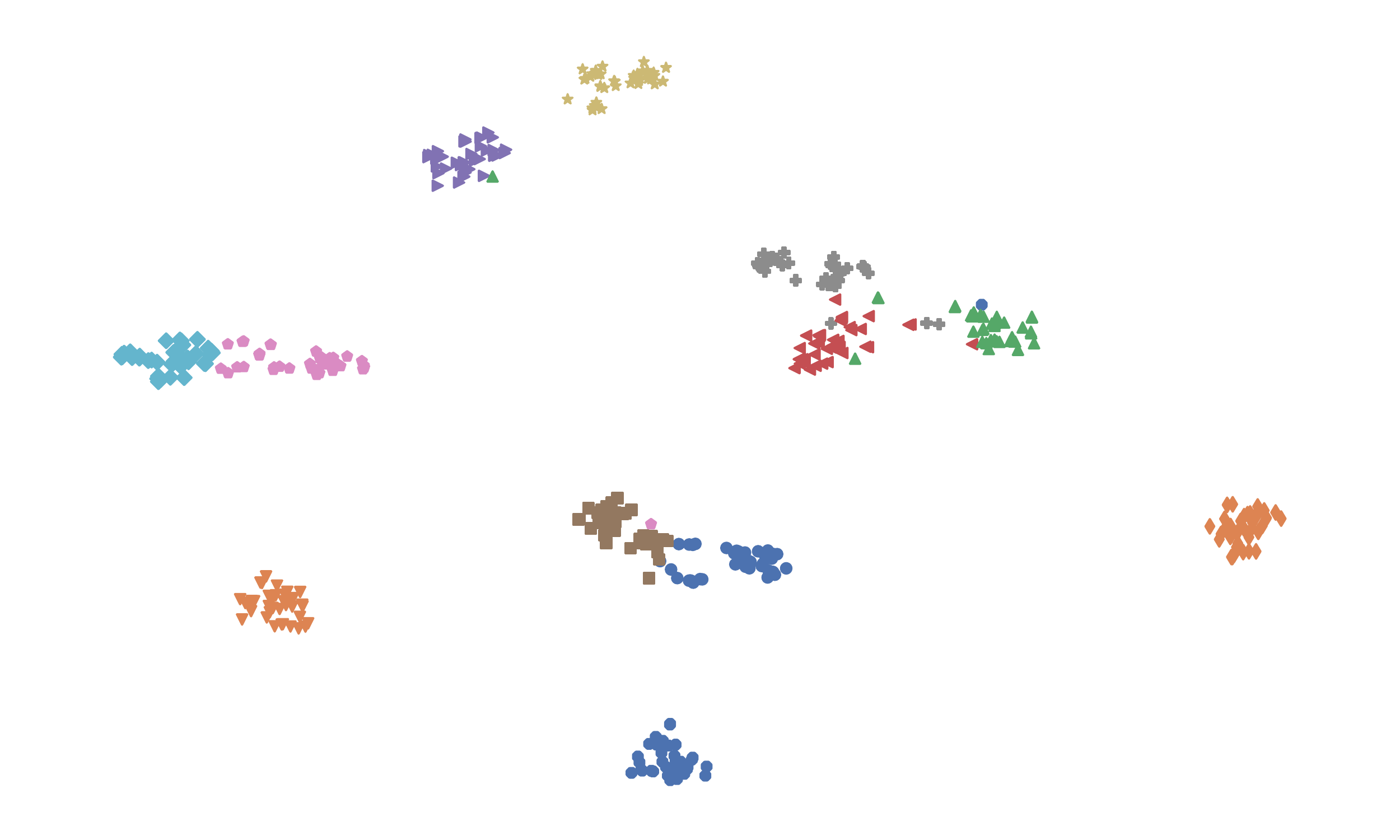}
        \caption{RoBERTa (30-shot S2)}
        \label{fig:rob-50k-10}
    \end{subfigure}
    \begin{subfigure}[!ht]{0.323\linewidth}
        \centering
        \includegraphics[width=0.98\linewidth]{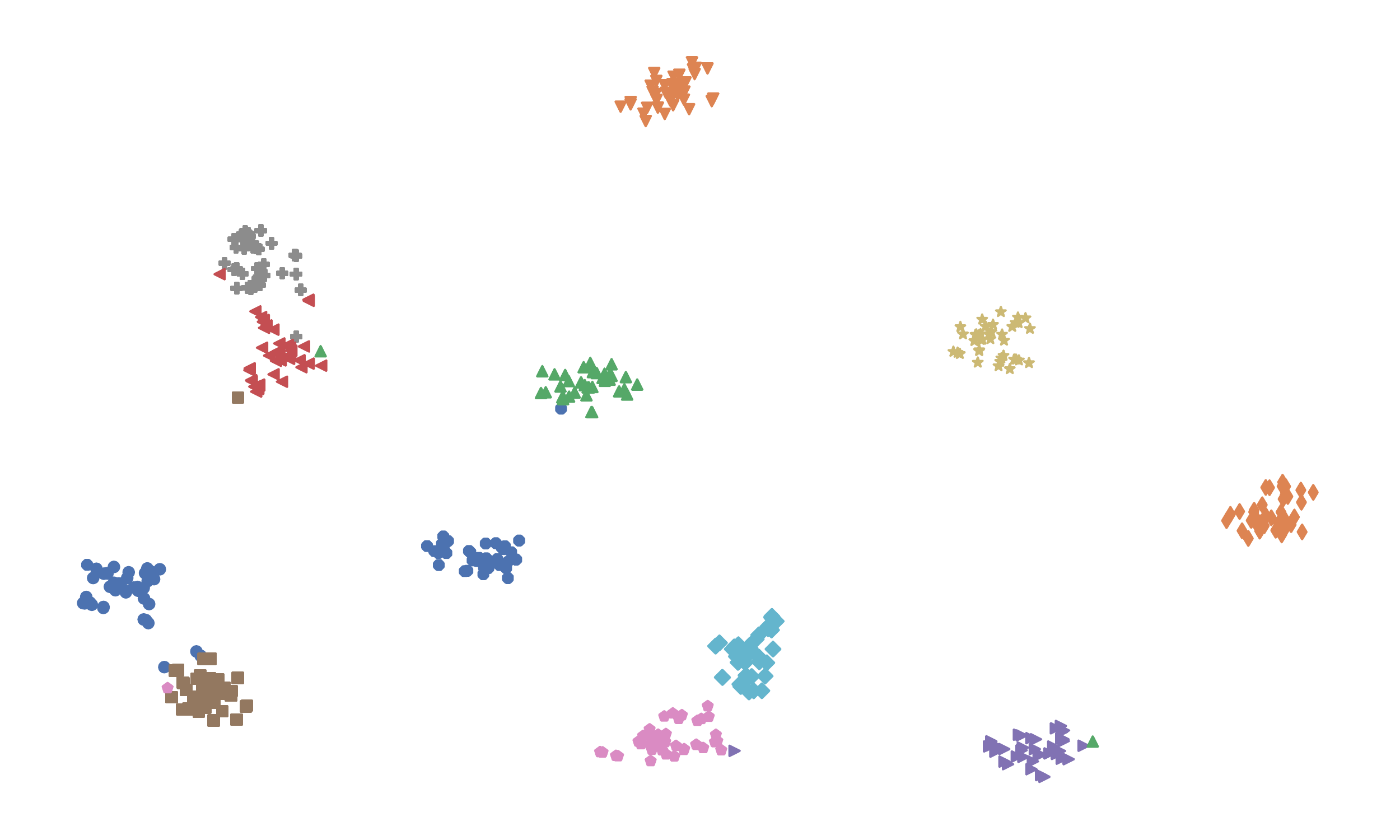}
        \caption{RoBERTa (Full S2)}
        \label{fig:rob-15m-10}
    \end{subfigure}
    \vspace{-1.5mm}
    \caption{t-SNE plots of encoded utterances from the test set of \textsc{banking77} (i.e., all examples are effectively unseen by the encoder models at training) associated with a selection of 12 intents. The encoded utterances are created via mean-pooling based on RoBERTa as the input LM: \textbf{(a)} without any Stage 1 fine-tuning with Reddit data; \textbf{(b)} Stage 1 fine-tuning with only 50k \textit{(context, response)} Reddit pairs; \textbf{(c)} Stage 1 fine-tuning with 2\% of the full Reddit corpus of \newcite{Henderson:2019arxiv} ($\approx$15M pairs). Stage 2 in all three cases is performed in 10-shot setups with the \nce objective ($n=3$ negatives). The respective Silhouette coefficient scores (higher is better): (a) $\sigma=0.320$, (b) $\sigma=0.338$, (c) $\sigma=0.378$.}
    \vspace{-1.5mm}
\label{fig:tsne-s1-10}
\end{figure*}

\begin{figure*}[t!]
    \centering
    \begin{subfigure}[!ht]{0.98\textwidth}
        \centering
        \includegraphics[width=1.0\linewidth,trim=0cm 0cm 0cm 0.1cm,clip]{./Figures/tsne-legend}
        \label{fig:legend3}
        \vspace{-3.5mm}
    \end{subfigure}
    \begin{subfigure}[!ht]{0.323\linewidth}
        \centering
        \includegraphics[width=0.98\linewidth]{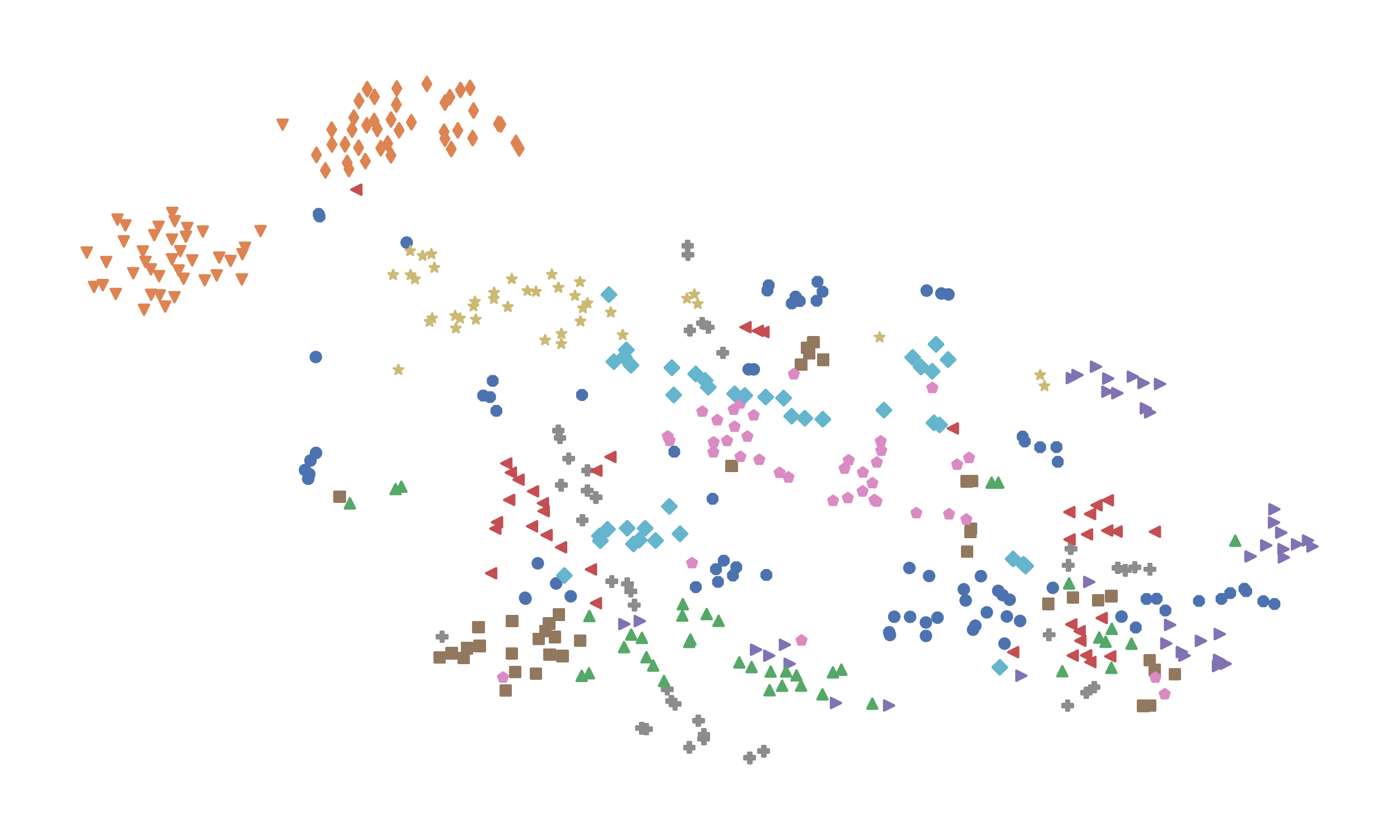}
        \caption{DistilRoBERTa (no fine-tuning)}
        \label{fig:tsne-droborig}
    \end{subfigure}
    \begin{subfigure}[!ht]{0.323\textwidth}
        \centering
        \includegraphics[width=0.98\linewidth]{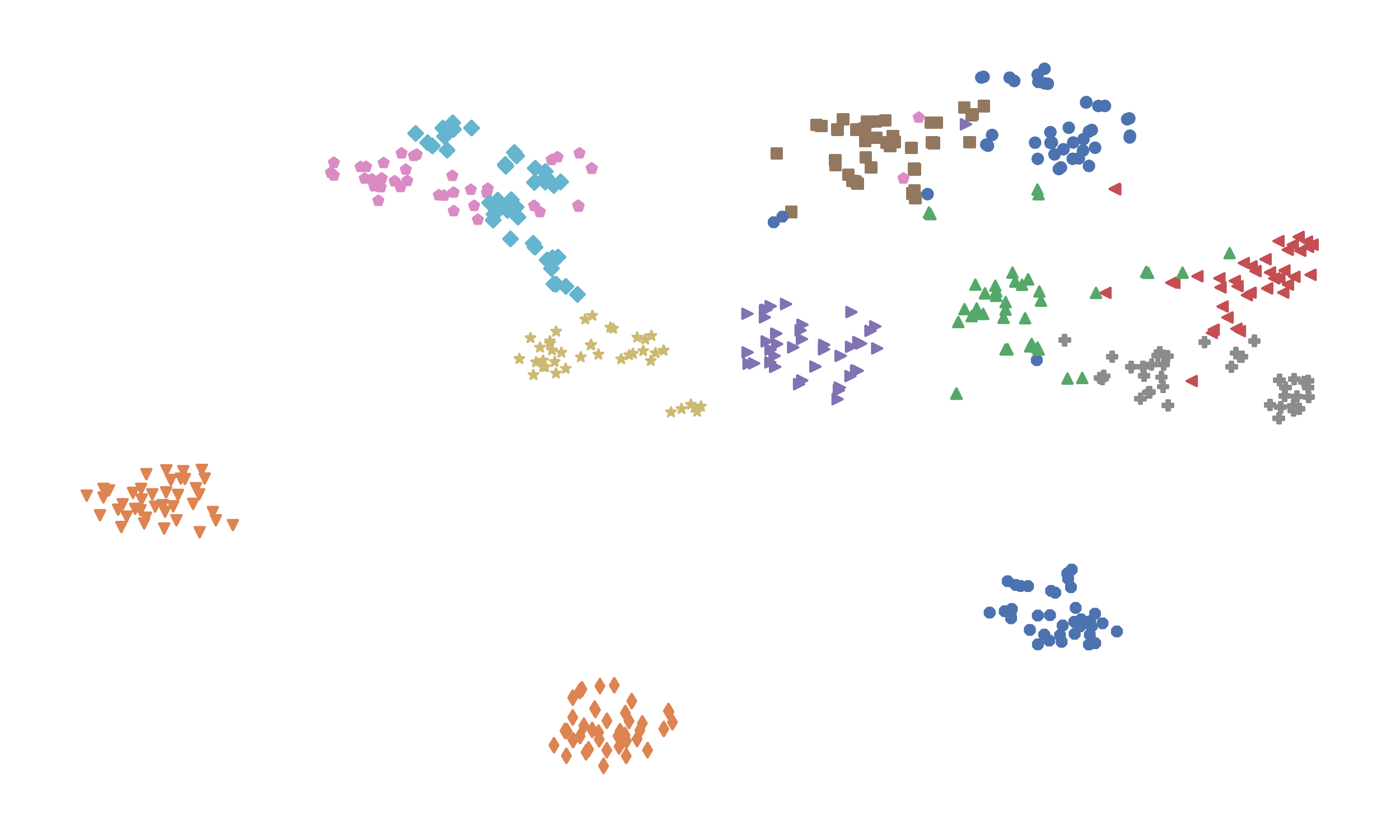}
        \caption{DistilRoBERTa (after S1)}
        \label{fig:tsne-drobr15m}
    \end{subfigure}
    \begin{subfigure}[!ht]{0.323\linewidth}
        \centering
        \includegraphics[width=0.98\linewidth]{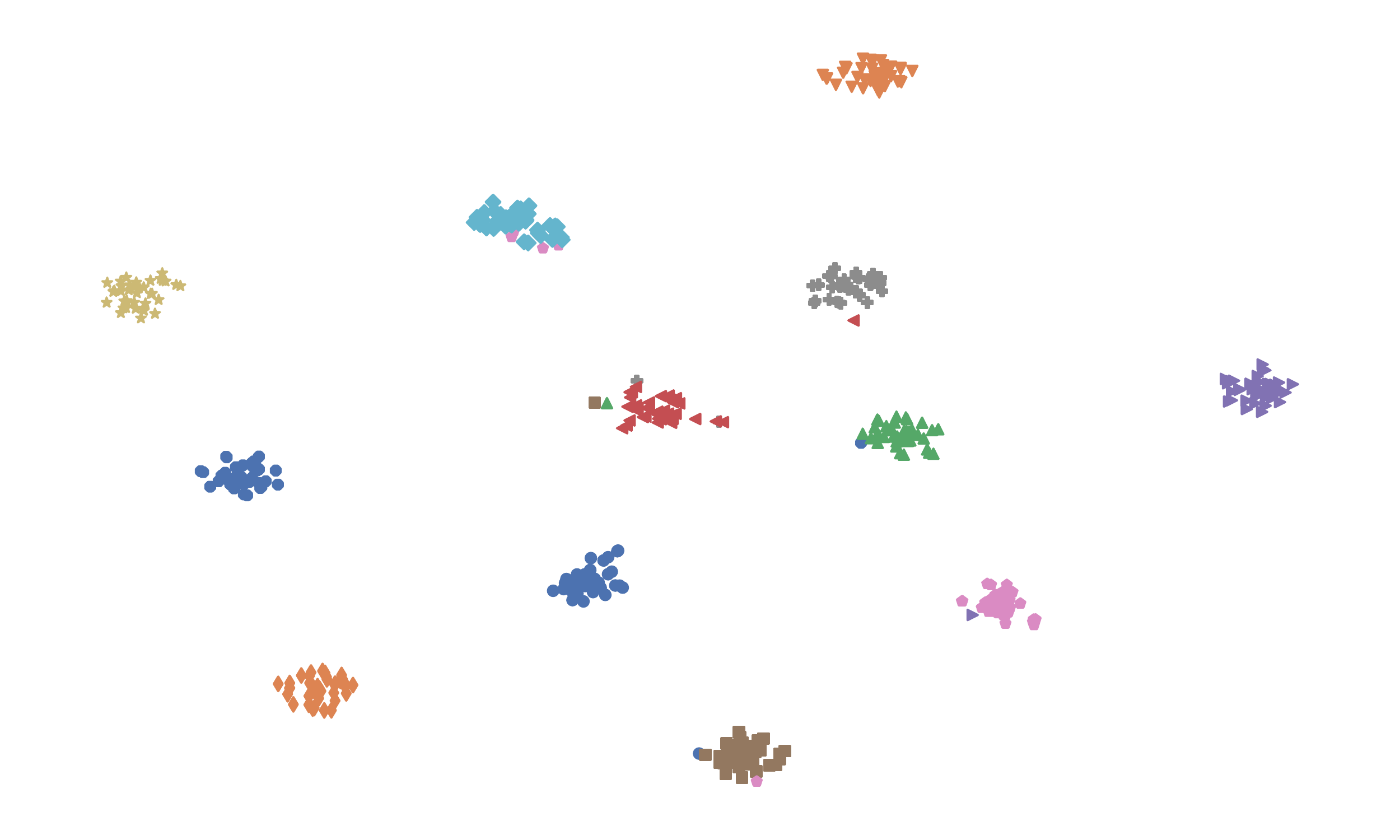}
        \caption{DistilRoBERTa (after S1 and S2)}
        \label{fig:tsne-drobcos}
    \end{subfigure}
    \vspace{-1.5mm}
    \caption{t-SNE plots of encoded utterances from the test set of \textsc{banking77} (i.e., all examples are effectively unseen by the encoder models) associated with a selection of 12 intents. The encoded utterances are created via mean-pooling based on \textbf{(a)} the original DistilRoBERTa LM; \textbf{(b)} DistilRoBERTa after Stage 1 (i.e., fine-tuned on 2\% of the full Reddit corpus, see Figure~\ref{fig:overview}); \textbf{(c)} DistilRoBERTa after Stage 1 and Stage 2, fine-tuned with the \cosine objective ($n=3$ negatives) using the entire \banking training set (see Figure~\ref{fig:overview}).}
\label{fig:tsne-drob}
\end{figure*}

\vspace{1.5mm}
\noindent \textbf{Figure~\ref{fig:tsne-1030-rob}} provides t-SNE plots with varying amounts of task data for Stage 2 task-tuning (10-shot versus 30-shot versus Full data setups), demonstrating that very tight and coherent clusters emerge even in the 10-shot setups. \textbf{Figure~\ref{fig:tsne-s1-10}} shows t-SNE plots after 10-shot Stage 2, when varying amounts of Reddit data for Stage 1 fine-tuning are used (e.g., skipping Stage 1 completely versus using $\approx$50k \textit{(context, response)} Reddit pairs). Finally, \textbf{Figure~\ref{fig:tsne-drob}} demonstrates that the patterns which emerge after Stage 1 and Stage 2 \convfit-ing do not depend on the chosen input LM, and on the chosen loss function in Stage 2: the trends very similar to Figure~\ref{fig:tsne-rob} (provided in the main paper) are also observed with \textit{distilRoBERTa} as the input LM, and \cosine as the S2 loss. \textbf{Figure~\ref{fig:tsne-r50k}} shows visible impact of adaptive Stage 1 fine-tuning even when only $50k$ Reddit \textit{(context, response)} pairs are used.

\begin{table*}[p]
\def\arraystretch{0.93}
\centering
{\footnotesize
\begin{tabularx}{\linewidth}{l XXX XXX XXX}
\toprule
  {} & \multicolumn{3}{c}{\bf \textsc{banking77}} & \multicolumn{3}{c}{\bf \textsc{clinc150}} & \multicolumn{3}{c}{\bf \textsc{hwu64}} \\
  \cmidrule(lr){2-4} \cmidrule(lr){5-7} \cmidrule(lr){8-10}
\textbf{Model Variant} & \textbf{10}  & \textbf{30} & \textbf{Full} & \textbf{10}  & \textbf{30} & \textbf{Full} & \textbf{10}  & \textbf{30} & \textbf{Full} \\
\cmidrule(lr){2-10}
 \cmidrule(lr){1-10}
 \rowcolor{Gray}
 {} & \multicolumn{9}{c}{\bf Similarity-Based Classification} \\
  \cmidrule(lr){2-10}
 \textsc{rob+s1+s2-cos} & \underline{86.48} & \underline{91.33} & {\bf 94.35} & \underline{92.87} & \underline{95.91} & \underline{97.20} & {85.06} & {\bf 90.46} & {\bf 92.98} \\
 \textsc{bert+s1+s2-cos} & {84.32} & {90.91} & {93.91} & {91.80} & {95.58} & {96.56} & \underline{85.13} & {89.41} & {91.93} \\
 \textsc{drob+s1+s2-cos} & {85.13} & {90.75} & {94.06} & {91.64} & {95.48} & {97.00} & {83.64} & {89.68} & \underline{92.94} \\
 \hdashline
 \textsc{rand+s1+s2-cos} & {79.03} & {87.37} & {91.69} & {83.96} & {89.98} & {94.12} & {76.30} & {82.62} & {88.20} \\
 \cmidrule(lr){2-4} \cmidrule(lr){5-7} \cmidrule(lr){8-10}
 \textsc{rob+s1+s2-ocl} & {\bf 87.38} & {\bf 91.36} & {94.16} & {\bf 92.89} & {\bf 96.42} & {\bf 97.34} & {\bf 85.32} & \underline{90.06} & {92.42} \\
 \textsc{bert+s1+s2-ocl} & {85.97} & {90.65} & {93.77} & {91.53} & {95.53} & {96.82} & {85.04} & {89.41} & {92.21} \\
 \textsc{drob+s1+s2-ocl} & {86.04} & {90.78} & {93.89} & {91.98} & {95.60} & {97.04} & {83.64} & {89.50} & {92.84} \\
 \hdashline
 \textsc{rand+s1+s2-ocl} & {80.62} & {87.01} & {91.49} & {84.91} & {90.98} & {94.44} & {77.23} & {82.99} & {88.85} \\
 \cmidrule(lr){2-4} \cmidrule(lr){5-7} \cmidrule(lr){8-10}
  \textsc{rob+s2-cos} & {84.96} & {90.81} & \underline{94.19} & {91.56} & {95.64} & {96.78} & {84.52} & {89.87} & {92.19} \\
 \textsc{bert+s2-cos} & {81.27} & {90.32} & {93.73} & {89.58} & {95.08} & {96.54} & {82.90} & {89.12} & {91.78} \\
 \textsc{drob+s2-cos} & {83.28} & {90.58} & {93.91} & {89.47} & {95.32} & {86.78} & {82.43} & {89.41} & {92.10} \\
 \hdashline
 \textsc{rand+s2-cos} & {70.32} & {84.16} & {90.75} & {76.31} & {86.69} & {91.76} & {65.89} & {79.18} & {86.43} \\
 \cmidrule(lr){2-4} \cmidrule(lr){5-7} \cmidrule(lr){8-10}
 \textsc{rob+s2-ocl} & {85.78} & {90.98} & {93.77} & {92.64} & {95.40} & {96.87} & {84.76} & {89.31} & {92.01} \\
 \textsc{bert+s2-ocl} & {82.28} & {89.77} & {93.54} & {90.71} & {95.07} & {96.62} & {83.09} & {88.94} & {92.57} \\
 \textsc{drob+s2-ocl} & {82.60} & {90.65} & {93.38} & {90.78} & {95.02} & {96.69} & {81.69} & {88.75} & {92.38} \\
 \hdashline
 \textsc{rand+s2-ocl} & {63.15} & {81.30} & {89.71} & {69.91} & {85.53} & {92.18} & {60.48} & {76.67} & {86.90} \\
 \cmidrule(lr){2-4} \cmidrule(lr){5-7} \cmidrule(lr){8-10}
  \textsc{rob+s1+s2-smax} & {86.27} & {90.58} & {94.06} & {92.44} & {95.62} & {96.76} & {85.87} & {88.83} & {92.48} \\
 \textsc{bert+s1+s2-smax} & {84.44} & {90.16} & {93.09} & {90.31} & {93.84} & {95.91} & {83.28} & {88.18} & {92.29} \\
 \textsc{drob+s1+s2-smax} & {83.32} & {89.85} & {93.47} & {90.42} & {94.13} & {96.47} & {83.36} & {88.75} & {92.57} \\
 \hdashline
 \textsc{rand+s1+s2-smax} & {76.79} & {85.55} & {90.97} & {82.22} & {87.69} & {92.91} & {76.30} & {81.51} & {88.85} \\
 \cmidrule(lr){2-4} \cmidrule(lr){5-7} \cmidrule(lr){8-10}
 \textsc{rob+s2-smax} & {84.61} & {90.49} & {93.66} & {91.89} & {95.17} & {96.71} & {83.46} & {88.57} & {92.57} \\
 \textsc{bert+s2-smax} & {81.33} & {89.44} & {92.63} & {89.69} & {93.38} & {96.12} & {81.51} & {87.83} & {91.58} \\
 \textsc{drob+s2-smax} & {82.60} & {89.31} & {93.54} & {89.44} & {93.96} & {96.04} & {82.53} & {87.36} & {91.91} \\
 \hdashline
 \textsc{rand+s2-smax} & {73.38} & {83.67} & {90.32} & {76.71} & {85.53} & {92.62} & {68.77} & {79.74} & {88.94} \\
 \cmidrule(lr){2-10}
 \rowcolor{Gray}
 {} & \multicolumn{9}{c}{\bf Baselines: MLP Classification} \\
\cmidrule(lr){2-10}
\textsc{rob+s1} & {83.08} & {90.16} & {93.38} & {90.98} & {94.12} & {96.42} & {81.13} & {87.73} & {91.44} \\
\textsc{bert+s1} & {82.69} & {89.82} & {93.67} & {89.88} & {94.07} & {96.33} & {82.25} & {88.01} & {91.12} \\
\textsc{drob+s1} & {82.95} & {89.55} & {93.34} & {89.76} & {93.46} & {96.02} & {81.23} & {87.64} & {90.91} \\
\cmidrule(lr){2-4} \cmidrule(lr){5-7} \cmidrule(lr){8-10}
\textsc{ConveRT$^*$} & {83.32} & {89.37} & {93.01} & {92.62} & {95.78} & {97.16} & {82.65} & {87.88} & {91.24} \\
\textsc{USE$^*$} & {84.23} & {89.74} & {92.81} & {90.85} & {93.98} & {95.06} & {83.75} & {89.03} & {91.25} \\
\textsc{USE} (ours) & {82.95} & {89.09} & {92.81} & {90.27} & {93.54} & {94.91} & {82.71} & {88.20} & {91.64} \\
\textsc{LaBSE} & {81.69} & {88.96} & {92.60} & {90.89} & {93.41} & {95.12} & {81.60} & {86.15} & {90.99} \\
\cmidrule(lr){2-10}
 \rowcolor{Gray}
 {} & \multicolumn{9}{c}{\bf Baselines: Full Fine-Tuning} \\
\cmidrule(lr){2-10}
\textsc{\textsc{bert} (base)}$^{**}$ & {79.87} & {--} & {93.02} & {89.52} & {--} & {95.93} & {81.69} & {--} & {89.97} \\
\bottomrule
\end{tabularx}
}%
\caption{Accuracy scores ($\times$100\%) on the three intent detection data sets with varying number of training examples (\textbf{10} examples per intent; \textbf{30} examples per intent; \textbf{Full} training data). As mentioned in \S\ref{s:experimental}, $n=3$ negatives are used in Stage 2 for 10-shot and 30-shot setups, $n=1$ for the Full setup. The peak scores per column are in bold, the second best is underlined. *The scores were taken directly from prior work, and computed on different 10/30-shot samples (and are thus not directly comparable, \citealt{Zhao:2020arxiv}) **The scores achieved by full (regular) fine-tuning of \textsc{bert (base)} have been taken directly from \newcite{Mehri:2020dialoglue}, and were not available for the 30-shot setup.}
\label{tab:main-app}
\end{table*}



\begin{table*}[t]
\def\arraystretch{0.84}
\centering
{\footnotesize
\begin{tabularx}{\linewidth}{l XX XX XX}
\toprule
  {} & \multicolumn{2}{c}{\bf \textsc{banking77}} & \multicolumn{2}{c}{\bf \textsc{clinc150}} & \multicolumn{2}{c}{\bf \textsc{hwu64}} \\
  \cmidrule(lr){2-3} \cmidrule(lr){4-5} \cmidrule(lr){6-7}
\textbf{After} & \textbf{10}  & \textbf{30}  & \textbf{10}  & \textbf{30}  & \textbf{10}  & \textbf{30}  \\
 \cmidrule(lr){2-3} \cmidrule(lr){4-5} \cmidrule(lr){6-7}
 {Epoch 1} & {86.30} & {91.40} & {92.80} & {96.02} & {\bf 86.15} & {\bf 90.33}  \\
 {Epoch 2} & {\bf 87.38} & {91.36} & {92.89} & {\bf 96.42} & {85.32} & {90.06} \\
 {Epoch 5} & {87.28} & {\bf 91.46} & {\bf 93.29} & {96.32} & {85.69} & {89.98} \\

\bottomrule
\end{tabularx}
}%
\vspace{-1mm}
\caption{Impact of longer Stage 2 \convfit-ing on the final performance; \textsc{rob+s1+s2-ocl}.}
\label{tab:longer}
\vspace{-1mm}
\end{table*}

\section{Models and Evaluation Data}
URLs to the models are provided in Table~\ref{tab:models}. The intent detection evaluation data is available online:

\vspace{1mm}
\noindent 1. \banking, \clinc, and \hwu intent detection data have been downloaded from the DialoGLUE repository: \\
\url{github.com/alexa/dialoglue}

We use the 10-shot data provided in the repository, and use their script to generate 30-shot setups for all three datasets.

\vspace{1mm}
\noindent 2. The English ATIS intent detection dataset is extracted from the recently published MultiATIS++ dataset \cite{Xu:2020emnlp}, available here: \\
\url{github.com/amazon-research/multiatis}

For reproducibility, we will release the generated 10-shot and 30-shot data splits.
\vspace{1.5mm}

Our code is based on PyTorch, and relies on the two following widely used repositories:
\begin{itemize}
    \item \texttt{sentence-transformers} \\
    \url{www.sbert.net}
    \item \url{huggingface.co/transformers/}
\end{itemize}

\begin{table*}[!t]
\def\arraystretch{0.99}
\centering
{\footnotesize
\begin{tabularx}{\textwidth}{ll X}
\toprule
{\bf Name} & {\bf Abbreviation} & {\bf URL} \\
\cmidrule(lr){2-2} \cmidrule(lr){3-3}
{bert-base-cased} & {\textsc{bert}} & {\url{huggingface.co/bert-base-uncased}} \\
{roberta-base} & {\textsc{rob}} & {\url{huggingface.co/roberta-base}} \\
{distilroberta-base} & {\textsc{drob}} & {\url{huggingface.co/distilroberta-base}} \\
{LaBSE} & {LaBSE} & {\url{huggingface.co/sentence-transformers/LaBSE}} \\
{multilingual USE} & {USE} & {{\tiny \url{tfhub.dev/google/universal-sentence-encoder-multilingual-large/3}}} \\

\bottomrule
\end{tabularx}
}
\vspace{-1.5mm}
\caption{URLs of the language models used in this work.}
\label{tab:models}
\end{table*}